\documentclass[letterpaper]{article}
\usepackage{hyperref}
\usepackage{aaai}
\usepackage{times}
\usepackage{helvet}
\usepackage{courier}
\usepackage{array}
\usepackage{amssymb}
\usepackage{graphicx}
\usepackage{graphics}
\usepackage{amsmath}
\usepackage{adjustbox}
\usepackage{url}
\usepackage{booktabs}
\usepackage{amsthm}
\usepackage{parskip}
\usepackage{todonotes}
\usepackage{tabularx}
\usepackage{siunitx}
\usepackage{array}

\usepackage [autostyle, english = american]{csquotes}
\MakeOuterQuote{"}

\usepackage{natbib}

\begin{document}
%
\title{The Use of AI for Thermal Emotion Recognition: \\
A Review of Problems and Limitations in Standard Design and Data}

\author{Catherine Ordun\\Department of Information Systems\\University of Maryland \\Baltimore County\\ Booz Allen Hamilton \\
cordun1@umbc.edu
\And
Edward Raff\\ Booz Allen Hamilton \\ Department of Computer Science\\University of Maryland \\Baltimore County \\ raff\_edward@bah.com
\And 
Sanjay Purushotham\\Department of Information Systems\\University of Maryland \\Baltimore County\\
psanjay@umbc.edu}

\maketitle
\begin{abstract}
\begin{quote}
With the increased attention on thermal imagery for Covid-19 screening, the public sector may believe there are new opportunities to exploit thermal as a modality for computer vision and AI. Thermal physiology research has been ongoing since the late nineties. This research lies at the intersections of medicine, psychology, machine learning, optics, and affective computing. We will review the known factors of thermal vs. RGB imaging  for facial emotion recognition. But we also propose that thermal imagery may provide a semi-anonymous modality for computer vision, over RGB, which has been plagued by misuse in facial recognition. However, the transition to adopting thermal imagery as a source for any human-centered AI task is not easy and relies on the availability of high fidelity data sources across multiple demographics and thorough validation. This paper takes the reader on a short review of machine learning in thermal FER and the limitations of collecting and developing thermal FER data for AI training. Our motivation is to provide an introductory overview into recent advances for thermal FER and stimulate conversation about the limitations in current datasets. 

\end{quote}
\end{abstract}

\section{Introduction}
Computer vision algorithms that use data from the visible spectrum (e.g. RGB) face a variety of challenges when it comes to human Facial Emotion Recognition (FER) due to the representation of superficial facial features laying on the epidermis. Physiological response from stress, fatigue, or other stimuli cannot be visualized on RGB but can be visualized through thermal imagery due to the changes in temperature detected sub-cutaneously. Thermal image data that can capture temperature changes correlated to human vital signs can be a powerful set of data for telemedicine applications supporting healthcare providers as a diagnostic tool for assessing inflammation and stress \citep{kosonogov2017facial}. Skin temperature can correlate to certain vital signs and offers a non-invasive method to remotely assess patients. As the cost of high resolution thermal sensors decline and more researchers release thermal FER datasets, there is a great potential to apply thermal imagery for telemedicine purposes. Since the Covid-19 pandemic, governments around the world have begun using thermal sensors combined with AI tools for Covid temperature screening \citep{ting2020digital}. From the U.K, China, Italy, Australia, to the U.S., multiple companies are offering the promise of integrated thermal sensing with facial recognition (FR) \citep{vannatta}. We believe that with broader adoption of thermal FR due to changes in HIPAA rules due to Covid-19, it will only be natural that researchers will want to advance their technology towards emotion screening. We caution that before leaping to thermal FER, researchers should be fully aware of the restrictions and limitations of thermal imagery and the problems that may underlie existing thermal FER databases. The adoption of thermal imagery as a source for any human-centered AI task is not easy. Thus, the goal of this paper is to present the state of the literature and discuss the challenges hindering the full adoption of AI as a tool for thermal FER.

\begin{figure}[t!]
\centering
\includegraphics[width=0.30\textwidth]{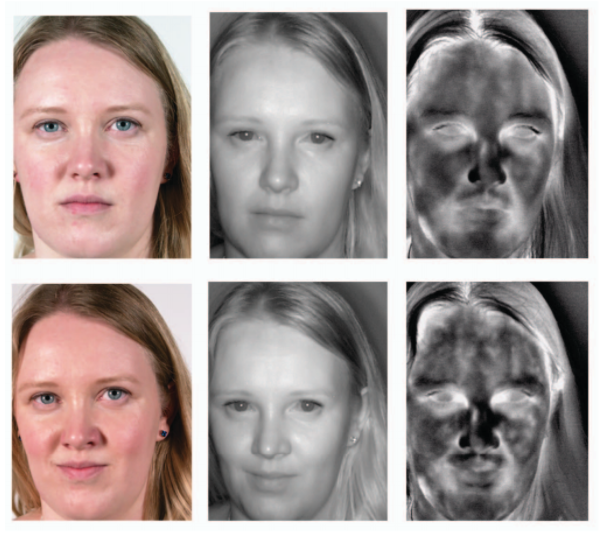}
\caption{RGB, near infrared and thermal images of a resting (up) and fatigued (down) face. In the thermal images, darker pixels corresponds to colder and lighter to hotter. \citep{lopez2017detecting}}
\label{lopez}
\end{figure}

\section{Advantages of Thermal over Visible}
When the public sector thinks about FER and facial recognition (FR), the go-to modality is the visible spectrum usually encoded as RGB. RGB images have dominated the area of FER, indicative through a variety of well known facial databases used in AI.\footnote{CK+ , FER 2013 , FERET , EmotioNet, RECOLA, Affectiva-MIT Facial Expression Dataset, NovaEmotions, MultiPIE, McMaster Shoulder Pain, AffectNet, Aff-Wild2, the Japanese Female Facial Facial Expression database, and CASME II for microexpressions} But, FR using RGB databases has become a controversial area of computer science, requiring careful consideration of its flaws and innate assumptions within the data and how it is applied \citep{martinez2019important,buolamwini2018gender,greene2019better,singer_metz_2019,lohr_2018}. Beyond the original intended academic purposes, some RGB databases have been taken down in order to prevent industry FR training \citep{murgia_2019}. In the wake of Black Lives Matters protests in June 2020, Microsoft and IBM discontinued their development of FR, where Amazon invoked a one year moratorium on FR based on evidence of algorithmic discrimination against communities of color \citep{matsakis_2020}.  Of particular value to the public sector, is whether thermal imagery for FER affords any level of privacy protection and bias mitigation. The answer may stem from the separation of thermal imagery from other machine learning tasks, known to increase recognition and decrease anonymity.

\begin{figure}[ht]
\centering
\includegraphics[width=0.40\textwidth]{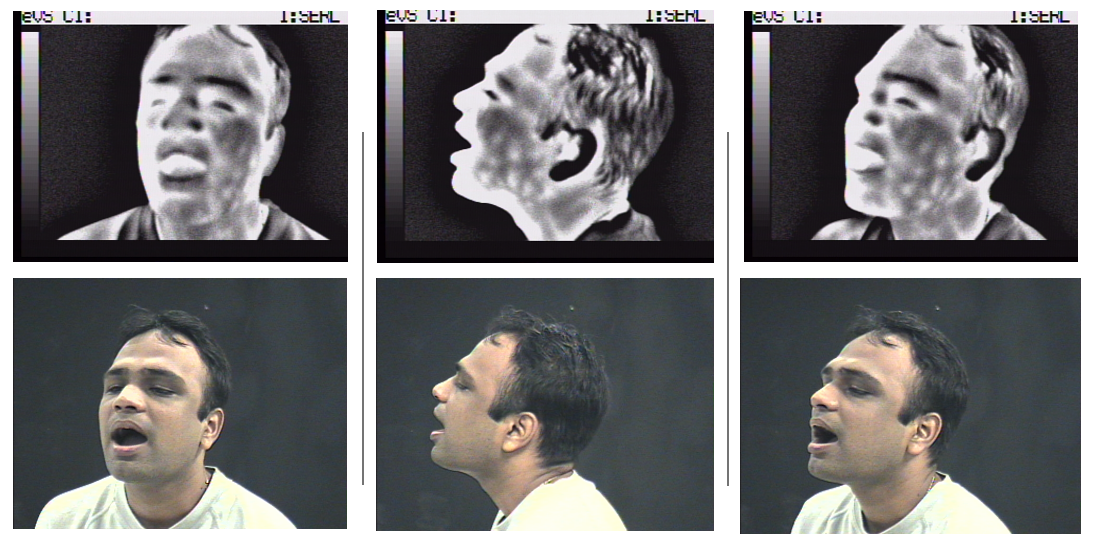}
\caption{Example of data from the Iris dataset \citep{iris}}
\label{iris}
\end{figure}

\begin{figure}[ht]
\centering
\includegraphics[width=0.40\textwidth]{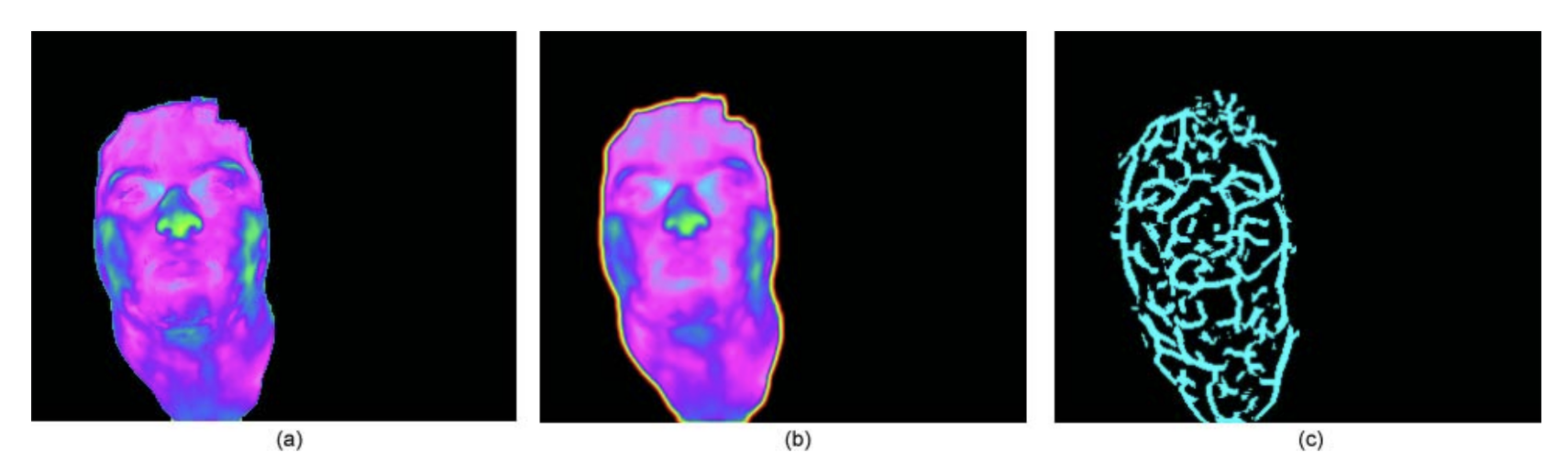}
\caption{Vascular network extraction: (a) Original segmented image; (b) Anisotropically diffused image; (c) Blood vessels extracted using white top hat segmentation, per \citep{buddharaju2007physiology}}
\label{budd}
\end{figure}

We believe that long-wave Infrared Radiation (LWIR) used alone, as a data source for FER, may be able to provide some form of anonymity for healthcare applications to minimize racial, ethnic, and potentially gender bias, when compared to RGB for FER. Through its low, grey-scale resolution \footnote{Thermal imaging manufacturers offer a variety of color palettes for visualizing temperature beyond "white hot" such as "iron bow" and "rainbow". It should be cautioned that some manufacturers offer fusion visualizations that fuse the RGB and thermal images together thereby improving resolution.} and reliance on temperature vectors driven by underlying vasculature \citep{ioannou2014thermal}, rather than superficial skin tone, texture, and pigmentation, thermal imagery can be more challenging to easily identify individuals. But there still remains a variety of issues to preserve privacy.
For example, anonymity may not be possible if thermal FER is combined with the machine learning task of FR, especially since thermal FR is well researched with multiple methods proposed to detect and recognize individuals. The concept of separating FR from other tasks is not uncommon. ~\citet{vannatta}, question whether during Covid-19 temperature monitoring, there is even a need to conduct FR given how the overall purpose is to identify infection as opposed to identity. It is important to caution, that although thermal FR is more challenging than the visible domain, it is feasible to use thermal imagery as a "soft" biometric due to its invariance under lighting and pose \citep{reid2013soft,friedrich2002seeing}. For example, superficial vascular networks are unique to each person's face as proposed by  \citet{buddharaju2007physiology}, and can be extracted through methods like anisotropic diffusion to identify minutiae points akin to fingerprints as shown in \autoref{budd}. Further, combining RGB with thermal can increase recognition accuracy. For example, \citet{nguyen2016body} used a combination of thermal and visible full body images for gender detection, finding that their proposed method of score-level fusion (training two separate SVM classifiers) combining thermal and visible led to a decrease in error of 14.672 equal error rate (EER) when compared to using thermal only (19.583 EER) and visible only (16.540 EER). 

\begin{figure}[ht]
\centering
\includegraphics[width=0.30\textwidth]{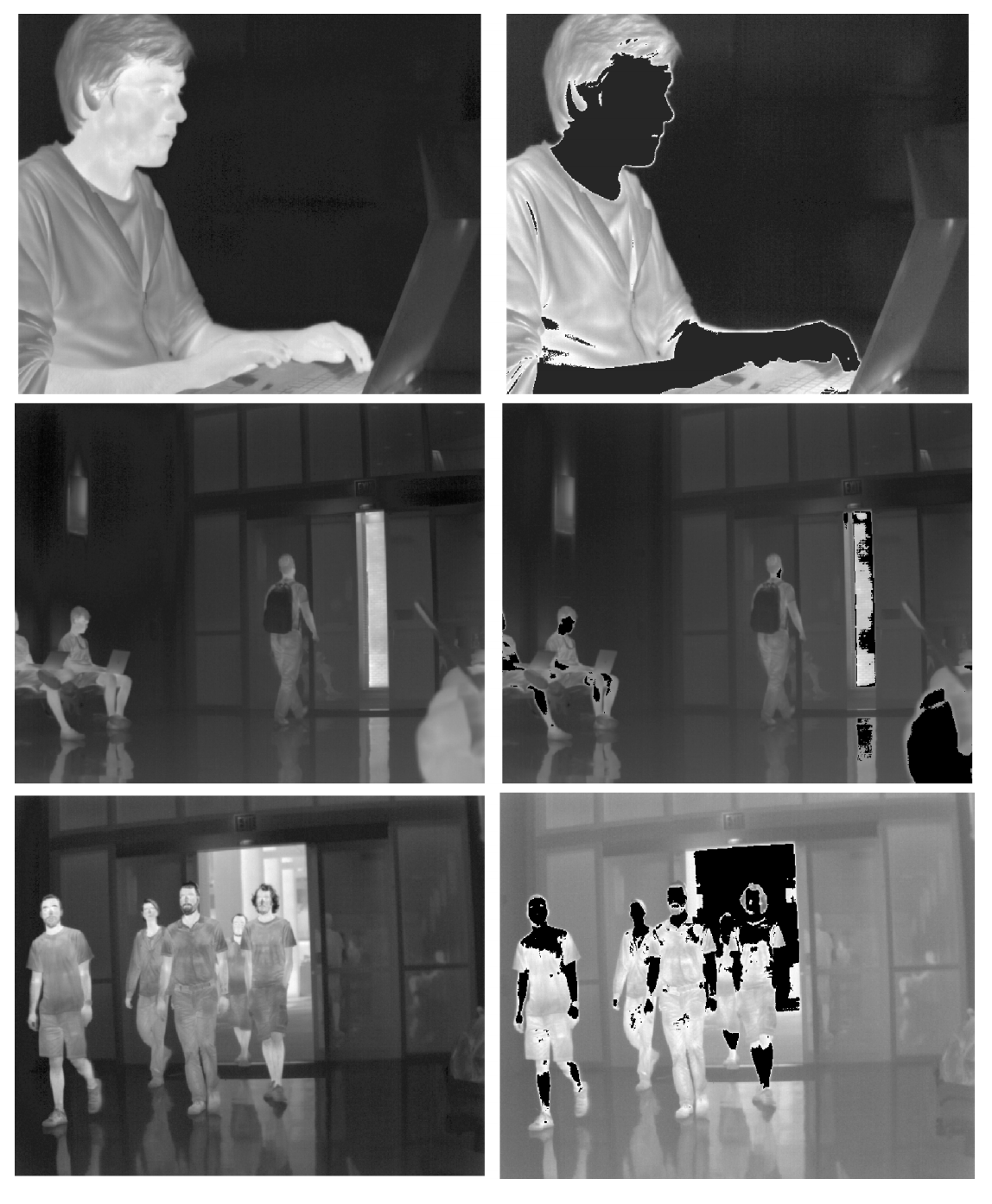}
\caption{Digitization privacy in different scenes: digitization results in scenes with people, computers and buildings. The left column are the input 16 bit images and the right column is the simulated output. \citep{pittaluga2016sensor}}
\label{dig}
\end{figure}

In addition, there has been research in the computer and electrical engineering fields to develop sensor-level privacy for thermal sensors in situations where people need to be sensed and tracked, but not identified. Work by \citet{pittaluga2016sensor} demonstrated different techniques to include digitization that masks human temperatures measurements thereby obscuring any ability to detect faces shown in \autoref{dig}, manipulating the sensor noise parameters as the thermal image is being generated, and algorithms to under or overexpose specific pixels that are designated as "no capture" zones. Still in research, these techniques require different levels of hardware and firmware upgrades based on the thermal sensor. 

Thermal imagery has additional technical advantages including how it is (1) invariant to lighting conditions unlike RGB, allowing the detection of physiological response (heat) to occur in low light or total darkness; (2) is a reliable and accurate correlation to standard physiological measures like respiration and heart rate; (3) is non-invasive i.e., requiring no skin contact whatsoever, making it convenient and non-intrusive and potentially relevant for non-communicative persons; (4) resistant to intentional deceit since physiological responses cannot be faked, whereas visible facial expressions can be controlled; and (5) is able to reveal facial disguises (i.e. wigs, masks) since these materials have high reflectivity and display as the brightest on thermograms compared to human skin which is among the darkest objects with low reflectivity \citep{pavlidis2000imaging}. In addition, thermal imagery offers physiological signals of social interactions from person to person. In terms of deceit detection, it is valuable to note that RGB images can also be used to detect microexpressions using databases like CASME II. Microexpressions are genuine, quick facial movements that may be uncontrollable or unnoticeable by the individual, and therefore have been studied as an indication of deception \citep{yan2014casme}. The RGB images used for studying microexpressions, however, are different than standard RGB FR datasets. They consist of video sequences captured using spontaneous natural elicitation, captured at a high frame rate of 200 fps, and labeled with facial action units (FAUs) which are encoded combination of facial movements based on Paul Ekman's Facial Action Coding System (FACS) \citep{ekman1999basic}.

\begin{table*}[]
\caption{Thermal Facial Emotion Recognition Datasets. }
\label{datasets}
\scalebox{0.80}{
\begin{tabular}{@{}lllllllllll@{}}
\toprule
Dataset             & Year & Pose   & Pairs & Affect & Subj & Access & Seq & Multi & THR         & VIS \\ \midrule
Univ. Notre Dame \citep{und}     & 2002 & Spont. & Yes   & UNK    & 241  & R      & UNK & Yes   & LWIR        & Yes \\
Equinox \citep{equinox,heo2004fusion}              & 2004 & Posed    & UNK   & 3      & 90   & N/A    & No  & No    & MW, LWIR    & Yes \\
IIT Delhi \citep{iit}           & 2007 & Posed  & UNK   & UNK    & 108  & R      & No  & UNK   & NIR         & No  \\
Univ. Houston \citep{buddharaju2007physiology}         & 2007 & Both   & Yes   & 0      & 138  & UNK    & UNK & No    & MWIR        & Yes \\
SC-Face  \citep{scface}             & 2009 & None   & Yes   & 0      & 130  & R      & No  & No    & NIR         & Yes \\
USTC-NVIE \citep{wang2010natural}           & 2010 & Both   & UNK   & 6      & 100  & N/A    & Yes & No    & LWIR        & Yes \\
Zhang \citep{zhang2010directional}                & 2010 & Posed  & UNK   & 0      & 350  & R      & No  & UNK   & NIR         & No  \\

UCHThermalFace \citep{hermosilla2012comparative}     & 2012 & Posed  & No    & 3      & 102  & UNK    & Yes & No    & LWIR        & UNK \\
KTFE Database \citep{nguyen2013thermal}        & 2013 & Spont. & Yes   & 7      & 26   & UNK    & Yes & No    & LWIR        & Yes \\
Iris \citep{iris}                 & 2013 & Posed  & Yes   & 3      & 30   & P      & No  & No    & LWIR        & Yes \\
RGB-D-T \citep{simon2016improved}            & 2016 & Posed  & Yes   & 5      & 51   & UNK    & UNK & UNK   & LWIR        & Yes \\
VIS-TH (Eurecom) \citep{mallat2018benchmark}     & 2018 & Posed  & Yes   & 4      & 50   & R      & Yes & Yes    & LWIR        & Yes \\
RWTH Aachen Univ. \citep{kopaczka2018fully}    & 2018 & Posed  & No    & 8      & 90   & R      & Yes & UNK   & LWIR        & No  \\
Tufts Face Database \citep{panetta2018comprehensive} & 2018 & Posed  & Yes   & 5      & 113  & R      & Yes & No    & NIR, LWIR   & Yes \\
UL-FMTV \citep{ghiass2014infrared}             & 2018 & Posed  & Yes   & UNK    & 238  & R      & Yes & Yes   & N, MW, LWIR & No  \\
ThermalWorld \citep{kniaz2018thermalgan}         & 2019 & Spont. & Yes   & 0      & 516  & R      & No  & No    & LWIR        & Yes \\
RFLDDJ \citep{seo2019face}              & 2019 & UNK    & Yes   & UNK    & UNK  & P      & UNK & No    & LWIR        & Yes \\ \bottomrule
\end{tabular}}
\small{Dataset - Database name, Year - publication year, Pose - Posed, Spontaneous, or Both, Pairs - Visible and Thermal, Affect - Number of labeled expressions, Subj - Number of unique human subjects, Access - R (requires permission from authors), P (publicly downloadable), Seq - Yes or No for availability in dataset of video sequences, Multi- Yes or No for multi-session recording, THR - Thermal image modality, VIS - Yes or No for presence of visible images, UNK means information was not provided in the paper}.
\end{table*}

\section{Physiology and Thermal FER}
A brief explanation of thermal radiation helps to understand how facial skin acts as a radiating surface. Thermal radiation is emitted by all objects above absolute zero (-273.15 $^{\circ}$C). Human skin is estimated at 0.98 to 0.99 $\epsilon$ \citep{yoshitomi2000effect}. The principal of thermal image generation is well understood by the Stefan-Boltzmann law that states total emitted radiation over time by a black body is proportional to $T^4$ where $T$ is temperature in Kelvins: $W = \epsilon \sigma T^4$ where $W$ is radiant emittance (${W}/{cm^2}$), $\epsilon$ is emissivity, $\sigma$ is the Stefan-Boltzmann constant ($5.6705 \cdot 10^{-12} {W}/{cm}^2 {K^4})$, and $T$ is Temperature ($K$).  

\begin{figure}[ht]
\centering
\includegraphics[width=0.40\textwidth]{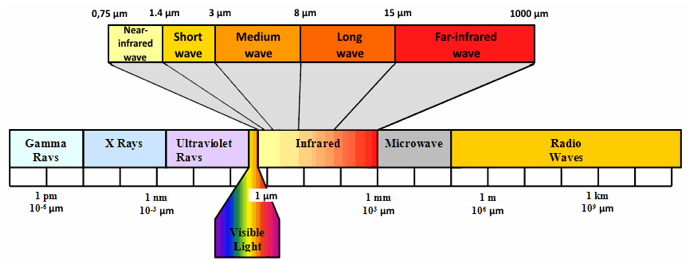}
\caption{Long-Wave IR falls in the wavelength range of \SI{8}{\micro\metre} to \SI{15}{\micro\metre}}
\label{spectrum}
\end{figure}

A black body is an object that absorbs all electromagnetic radiation it comes in contact with. No electromagnetic radiation passes through the black body and none is reflected. Since no visible light is reflected or transmitted, the object looks black upon visualization from thermal imagery, when it is cold. Thermal sensors respond to infrared radiation (IR) and produce visualizations of surface temperature.  Because LWIR operates in a sub-band of the electromagnetic spectrum per \autoref{spectrum} it is invariant to illuminating conditions meaning that it can operate in low light to complete darkness.  By imaging temperature variations to emotionally induced stimuli such as videos or pictures, thermograms reveal genuine responses to social situations. This occurs through activation of the autonomic nervous system (ANS) where emotional arousal leads to a perfusion of blood vessels innervated at the surface of the skin \citep{ioannou2014thermal}. 

\begin{figure}[ht]
\centering
\includegraphics[width=\columnwidth]{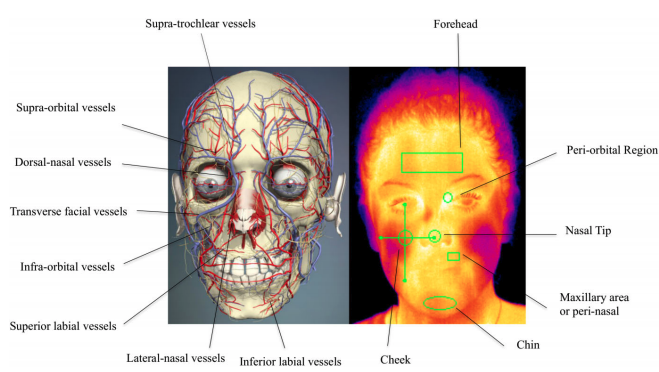}
\caption{Thermal representation for extraction of ROIs by Ioannou}
\label{ioa}
\end{figure}

These images are called thermograms and are the data captured in thermal FER datasets, with labels based of the emotional response elicited (i.e. happiness, disgust, sadness, deceit, stress, etc.). Although today's need for a touch-less system are paramount, the concept of using thermograms for contact-less physiological monitoring is not new and rooted in the intersection of physiological research (Selinger 2016;Buddharaju 2007;Pavlidis 2000; Ionnou 2014) and affective computing (Wilder 1996;Yoshitomi 2000;Goulart 2019). These include applications for FER where different emotions are detected from thermal facial images alone, in addition to person re-identification on thermal imagery, for FR.  Since 1996 \citep{wilder1996comparison} there have been numerous studies evaluating how thermograms correlate with vital measures. In 2007, Pavlidis~\citep{pavlidis2007interacting} demonstrated that thermal imagery is a reliable measure to assess emotional arousal where different regions of the face (zygomaticus, frontal, orbital, buccal, oral, nasal) correlate with different emotional responses.  Thermal imagery also visualizes the physiology of perspiration~\citep{pavlidis2012fast,ebisch2012mother}, cutaneous and subcutaneous temperature variations~\citep{hahn2012hot,merla2004emotion}, blood flow~\citep{puri2005stresscam}, cardiac pulse~\citep{garbey2007contact}, and metabolic breathing patterns~\citep{pavlidis2012fast} and has been used to monitor heat stress and exertion~\citep{bourlai2012use}. The reliability of thermal temperature readings have been repeatedly shown to be consistent and correlate accurately with gold standard physiological measures of electrocardiography (ECG), piezoelectric thorax stripe for breathing monitoring, nasal thermistors, skin conductance, or galvanic skin response (GSR)~\citep{pavlidis2007interacting,sonkusare2019detecting}. 

\begin{figure}[ht]
\centering
\includegraphics[width=\columnwidth]{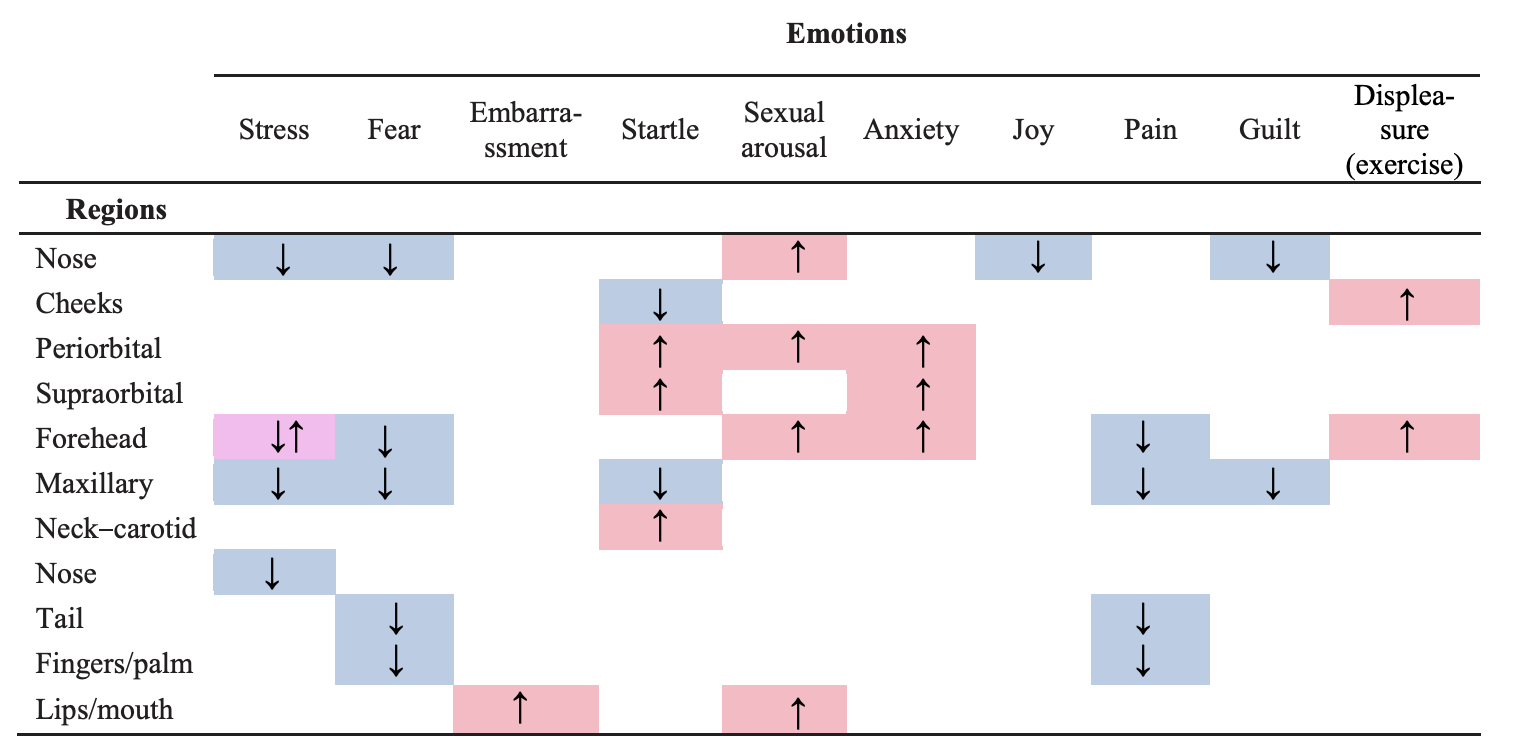}
\caption{Skin thermal variations in the considered regions of interest across emotions}
\label{rois}
\end{figure}

We can even observe these changes with the naked eye, such as embarrassment causing a person to blush \citep{sonkusare2019detecting}, or fear leading to pallor \citep{kosonogov2017facial}.  Merla~ \citep{merla2014revealing} offered a survey of thermal studies in psychophysiology from 1990 to 2013, demonstrating a series of emotional responses detected on thermal imagery such as startle response, fear of pain, lie detection, mental workload, empathy, and guilt.  These responses occur in different regions of the face, or ROIs. Salazar-Lopez found high arousal images elicited temperature increases on the tip of the nose~\citep{salazar2015mental}.  Kosnogov~\citep{kosonogov2017facial} found that more arousing an image, the faster and greater the thermal response on the tip of the nose. He speculated that the speed and magnitude of these thermal responses were linked to autonomic adjustments normal to emotional situations.  Zhu~\citep{zhu2007forehead} found that deception was detected through increased forehead temperature  and Puri~\citep{puri2005stresscam} found the forehead to be correlated with stress. Social responses based on one-on-one personal contact can also be observed. For example, Ebisch~\citep{ebisch2012mother} found "affective synchronization" of facial thermal responses between mother and child, where distress temperatures at the tip of the nose were mimicked by the mother as she watched her child in distress. Fernandez~\citep{fernandez2015classification} summarizes analysis by \citet{ioannou2014thermal} describing whether temperature increases, decreases, or stays the same based on different emotions and ROIs provided in \autoref{rois}.

\begin{table*}[]
\caption{Selected Thermal Facial Emotion Recognition AI Papers}
\label{aithermalpapers}
\adjustbox{max width=\textwidth}{
\begin{tabular}{@{}lllllllllll@{}}
\toprule
Author         & Year & Affect           & ROIs               & Model     & Dataset             & Target           & Acc        & Data & Code & Params \\ \midrule
Stemberger & 2010 & Cognitive Workload & 7 ROIs & ANN & Custom dataset & Multiple Workload & 81.0\% & (-) & (-) & (+) \\
Wang            & 2014 & Spont. Affect    & Whole face         & DBM       & USTC-NVIE           & Valence          & 62.9\%          & (+)     & (-)  & (+)    \\
Wu             & 2016 & Posed Affect     & Whole face         & CNN       & RGB-D-T             & Multiple Affects & 99.40\%         & (-)     & (-)  & (-)    \\
Simon         & 2016 & Posed Affect     & Whole face         & CNN       & RGB-D-T             & Multiple Affects & UNK             & (-)     & (-)  & (+)    \\
Cho           & 2017 & Stress           & Nose               & CNN       & Custom dataset      & Binary Stress    & 85.59\%         & (-)     & (-)  & (+)    \\
Lopez          & 2017 & Exercise Fatigue & Whole face, 3 ROIs & CNN, SVM  & Custom dataset      & Binary Fatigue   & 23.3\% - 81.8\% & (+)     & (-)  & (+)    \\
Haque & 2018 & Pain & Whole face & CNN, LSTM & Custom dataset & 5 Pain Levels & 18.33\% CNN & (+) & (-) & (+) \\
Ilyas         & 2018 & Spont. Affect  & Whole face         & CNN, LSTM & Custom dataset      & Multiple Affects   & 89.74\%         & (-)     & (-)  & (-)    \\
Elbarawy &  2019 & Posed Affect & Whole face & CNN & Iris & Multiple Affects & 96.7\% & (+) & (-) & (+) \\

Ilikci       & 2019 & Posed Affect     & Whole face         & CNN       & Iris                & Multiple Affects & 92.72\%         & (+)     & (-)  & (+)    \\
Shreyas Kamath  & 2019 & Posed Affect     & Whole face         & CNN       & Tufts Face Database & Multiple Affects & 96.2\%          & (+)     & (-)  & (+)    \\ \bottomrule
\end{tabular}}
\small{Year - Publication year, Affect - Expression type (Posed and Spont. mean basic discrete emotions), ROIs - facial regions of interest, Model - Deep learning algorithm type, Dataset - name of database, Target - the predicted class (all papers identified were classification), Acc - Best classification accuracy across models reported. Data - link to database provided if custom or name of public database provided, Code - link to code provided, Params - model parameters disclosed in paper, Annotations of (-) indicate information not disclosed, and (+) means it was disclosed in the paper}.
\end{table*}

\section{AI and Thermal FER}
Since 2000 with \citep{yoshitomi2000effect}, machine learning in thermal FER has grown slowly to include emotion classification by \citep{khan2006automated,nhan2009classifying,wang2014emotion,jarlier2011thermal,wang2014fusion,Trujillo2005automatic} with gradual adoption of AI methods such as neural networks. The ability to move away from manual, hand-crafted feature extraction to automatic learning through neural networks has already proven advantageous for thermal-to-visible image translation through GANs \citep{mallat2019cross,kniaz2018thermalgan,chen2019matching}, and for automated temperature vector extraction of facial ROIs \citep{sonkusare2019detecting}. Earlier works in deep learning applied to thermal FER such as the works of \citet{wang2014emotion} using a Deep Boltzman Machine (DBM) found that learning feature representations directly from thermal images of the NVIE dataset \citep{wang2010natural} led to greater accuracy (62.9\%) when predicting low and high valence, compared to statistical temperature vectors manually extracted from thermal images followed by dimensionality reduction (PCA) and SVM (61.0\%). Further, Wang asserted that the DBM learned from features representing a mixture of thermal datasets such as the Equinox \citep{equinox} and NVIE led to greater accuracy by 5.3\%.  Thermal features can outperform visible features in FER, overall, even without deep learning methods. Goulart's thermal multi-affect classifier using PCA and LDA outperformed visible emotion classifiers particularly on challenging expressions such as disgust and fear which can range between 40\% to 50\% for RGB accuracy. Whereas, Goulart's thermal classifiers detected disgust with 89.93\% and fear at 88.22\% true positive rates \citep{goulart2019emotion}.  

\begin{figure}[ht]
\centering
\includegraphics[width=\columnwidth]{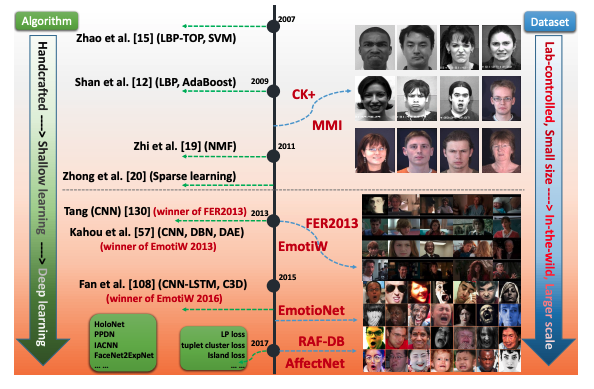}
\caption{Growth of lab-controlled, small size data to "in-the-wild", larger scale data encouraged use of deep learning algorithms in visible FER \citep{li2018deep}}
\label{Li}
\end{figure}

\citet{li2018deep} describes how AI research in the visible domain grew based on the broad dissemination of public, large-scaled, natural data per \autoref{Li}.  Any internet search will reveal dozens of RGB FR databases easily accessible and downloadable, such as the Top 15 list of facial recognition databases on Kaggle \citep{hamdi_2020}. They identified 74 visible "deep FER" papers using CNNs, Generative Adversarial Networks (GANs), Restricted Boltzman Machines (RBM), Deep Auto Encoders, Deep Belief Networks (DBN), and Recurrent Neural Networks (RNN) trained on such RGB FR datasets. But, AI in thermal FER lags behind, possibly due to the lack of large-scale, publicly available, and comprehensive thermal FER datasets.

\begin{table*}[]
\caption{Examples of Thermal FER Experimental Design Parameters}
\label{design}
\adjustbox{max width=\textwidth}{
\begin{tabular}{@{}llllllllllll@{}}
\toprule
Author &
  Year &
  Thermal Cam. &
  Dual Sensor &
  Thermal Res. &
  Dem. &
  Exclusion &
  Subjects &
  Temp. &
  Rest Time &
  Lighting &
  Stimulus \\ \midrule
Nhan &
  2010 &
  ThermaCAM &
  UNK &
  UNK &
  9F, 3M, mean 24 yo &
  UNK &
  12 &
  UNK &
  20 min &
  UNK &
  Static images \\
Wang  &
  2010 &
  SAT-HY6850 &
  UNK &
  320 x 240 &
  58F, 157M, 17 - 31 yo &
  UNK &
  215 &
  Means 23.29 &
  UNK &
  Yes &
  Emotional videos \\
Hermosilla  &
  2012 &
  Flir 320 TAU &
  UNK &
  324 x 256 &
  UNK &
  UNK &
  102 &
  UNK &
  UNK &
  UNK &
  UNK \\
Nguyen &
  2013 &
  NEC R300 &
  Yes &
  UNK &
  UNK gender, 11 - 32 yo &
  UNK &
  26 &
  24 - 26 &
  2 hrs. &
  UNK &
  Emotional videos \\
Salazar-Lopez &
  2015 &
  ThermoVision A320G &
  UNK &
  UNK &
  60F, 60M, 24 - 27 yo &
  Yes &
  120 &
  18 - 25 &
  10 - 15 min. &
  UNK &
  Static images \\
Lopez &
  2017 &
  Therm-App &
  UNK &
  288 x 384 &
  8F, 11M, 23 - 27yo &
  UNK &
  19 &
  UNK &
  Until heart rate below 20 bpm &
  UNK &
  Exercise \\
Mallat &
  2018 &
  Flir Duo R &
  Yes &
  160 x 120 &
  No &
  UNK &
  50 &
  25 &
  No &
  Yes &
  UNK \\
Goulart &
  2019 &
  Therm-App &
  UNK &
  384 x 288 &
  8F, 9M, 8 - 12 yo &
  UNK &
  17 &
  20 - 24 &
  10 min. &
  Yes &
  Questionairre \\
Sonkusare &
  2019 &
  Flir A615 &
  UNK &
  640 x 480 &
  11F, 9 M, 22 - 30 yo &
  Yes &
  20 &
  22 &
No alcohol \& caffeine 2 hrs. prior &
  Yes &
  Auditory stimulus \\
Panetta &
  2020 &
  FLIR Vue Pro &
  UNK &
  UNK &
  UNK &
  UNK &
  113 &
  UNK &
  UNK &
  Yes &
  UNK \\ \bottomrule
\end{tabular}}
\small{Year - Publication year,  Thermal Cam. - Type of LWIR camera, Dual Sensor - Yes or No, captures visible and thermal simultaneously, Thermal Res. - Reported thermal pixel resolution, Dem. - Demographics of subjects, Exclusion - Yes or No, exclusion or inclusion criteria documented, Subjects - Number of unique human subjects, Temp. - Room temperature for experiment reported in degrees Celsius, Rest Time - Time subjects reach relaxed state prior to image capture, Lighting - Yes or No, illumination design documented, Stimulus - Type of stimulus to provoke spontaneous response, if spontaneous, UNK means information was not found in the paper}.
\end{table*}

Where Li identified 74 papers, we only identified 14 thermal FER datasets in \autoref{datasets} whose numbers have increased since 2018 possibly due to the decreasing cost of thermal cameras and the easier ability to purchase them online.  Further, we identified only eleven AI thermal FER papers, shown in  \autoref{aithermalpapers}, starting in 2010, indicating a slow evolution from manual feature extraction using geometric methods to learning latent representations using deep learning. These works do not consistently release code and have varied levels of explanation around experimental design and arousal stimulus, which we summarized in \autoref{design}. This makes it challenging to reproduce, much less compare across studies.

\begin{figure}[t!]
\centering
\includegraphics[width=0.45\textwidth]{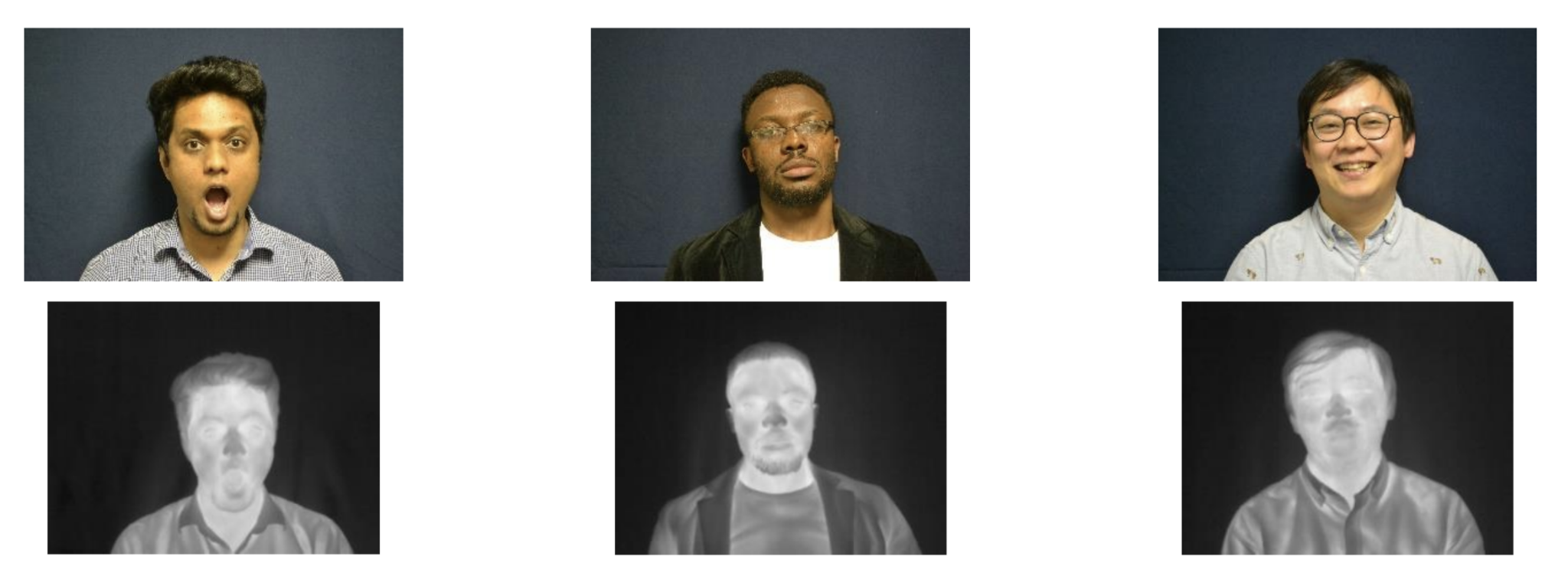}
\caption{The Tufts Face Database \citep{panetta2018comprehensive}}
\label{tufts}
\end{figure}

Researchers in thermal emotion recognition such as \citet{goulart2019emotion} agree, particularly since there is no standard thermal FER imaging benchmark dataset consistently used across studies. In an empirical review reproducing 255 machine learning papers, Raff~\citep{raff2019step} notes that papers which are scientifically sound and complete, should be independently reproducible based solely on explanation, details, and descriptions. Failures in reproducibility can occur when language or notation is unclear, when the algorithm is missing details about implementation or equations, and when nuanced details are left out. In~\autoref{datasets} we catalog the few available (via request or publicly) thermal datasets that have been used for tasks including FR and FER. They vary in scope, where some do not have emotion labels at all, making it difficult to benchmark and standardize results that may eventually impact psychological and health-related decisions. One example of a recently developed thermal FER dataset is by Tufts University shown in \autoref{tufts}.

\section{Thermal FER Data Challenges}
Some researchers have noticed the lack of variation across thermal FR dataset that fail to account for diverse emotional states, alcohol intake or exercise, and ambient temperature, leading them to doubt the rigor of the reported results especially in real life conditions \citep{shoja2014face}. Assuming that the lack of a comprehensive thermal FER benchmark dataset is one factor that hinders the advancement of AI research, we can begin exploring the challenges of designing such a dataset. But, developing a thermal FER dataset is different than simply crawling the web for RGB faces. The collection of thermal FER data requires an experiment unto itself, needing institutional review board (IRB) approval, subject recruitment, experimental design, and specialized equipment. As a result, thermal FER datasets are expensive in terms of time and labor. We have observed some trends across databases that if addressed in the development of a single high-fidelity dataset, may carve a path for greater adoption of thermal AI FER studies. We justify these assertions based on research in the psycho-physiology domain, below.

\subsection{Include video sequences}
Video sequences present timing of the arc of expression onset and delay. It is important to capture intensity and duration of expression which has been found consistent with automatic movement and neuropsychological models \citep{tian2005facial}. \citet{levenson1988emotion} indicated that duration of an emotional response is 0.5 – 4 seconds. But Nguyen~~\citep{nguyen2013thermal} cites mistakes in many of the leading thermal recognition databases. In the USTC-NVIE database their procedure for data acquisition had video gaps between each emotion clip at 1-2 minutes which is too short for participants to establish a neutral emotion status. Research indicates that for thermal response (cutaneous skin temperature), there is a delay after stimulus that needs to be accounted for and recorded \citep{ioannou2014thermal} and temperature change can occur in less than 30 seconds upon stimulation \citep{pavlidis2012fast}.  Temperature changes at the tip of the nose can occur as fast as 10 seconds after stimulus and last 20 - 30 seconds regardless of distress or soothing \citep{ebisch2012mother}. In a more recent paper, \citep{sonkusare2019detecting} were able to quantify the temporal dynamics of thermal response when compared to gold standard measures like Galvanic Skin Response (GSR) demonstrating that thermal response occurred only 2 seconds later than GSR when exposed to an auditory stimulus.  Static images without a time axis can be incomplete and will fail to capture the complete physiological signal and emotional response. 
\subsection{Enable spontaneous response}
Many existing thermal databases that are focused only on FR have discrete, posed affects based on the labeling defined by Ekman (Ekman 1999). But affective researchers argue that spontaneous emotional reactions are more realistic since, "people show blends of emotional displays…hence, the classification of human non-verbal affective feedback into a single “basic”-emotion category may not be realistic." \citep{gunes2010automatic,mcduff2017large}.  Further, multiple emotions typically occur as opposed to a single discrete response. For example, in a 1993 study by Gross et al. 85 subjects self-reported a variety of feelings after watching a close-up arm amputation medical video \citep{gross1993emotional}. 

\begin{figure}[ht]
\centering
\includegraphics[width=\columnwidth]{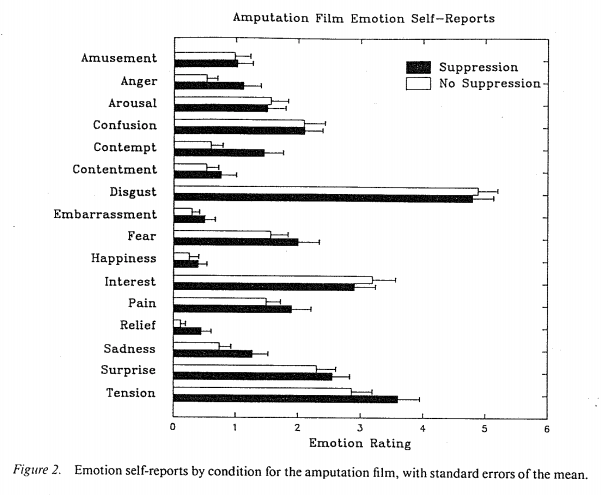}
\caption{Multiple feelings self-reported after exposure to high arousal video \citep{gross1993emotional}}
\label{gross}
\end{figure}

\begin{figure}[ht]
\centering
\includegraphics[width=0.40\textwidth]{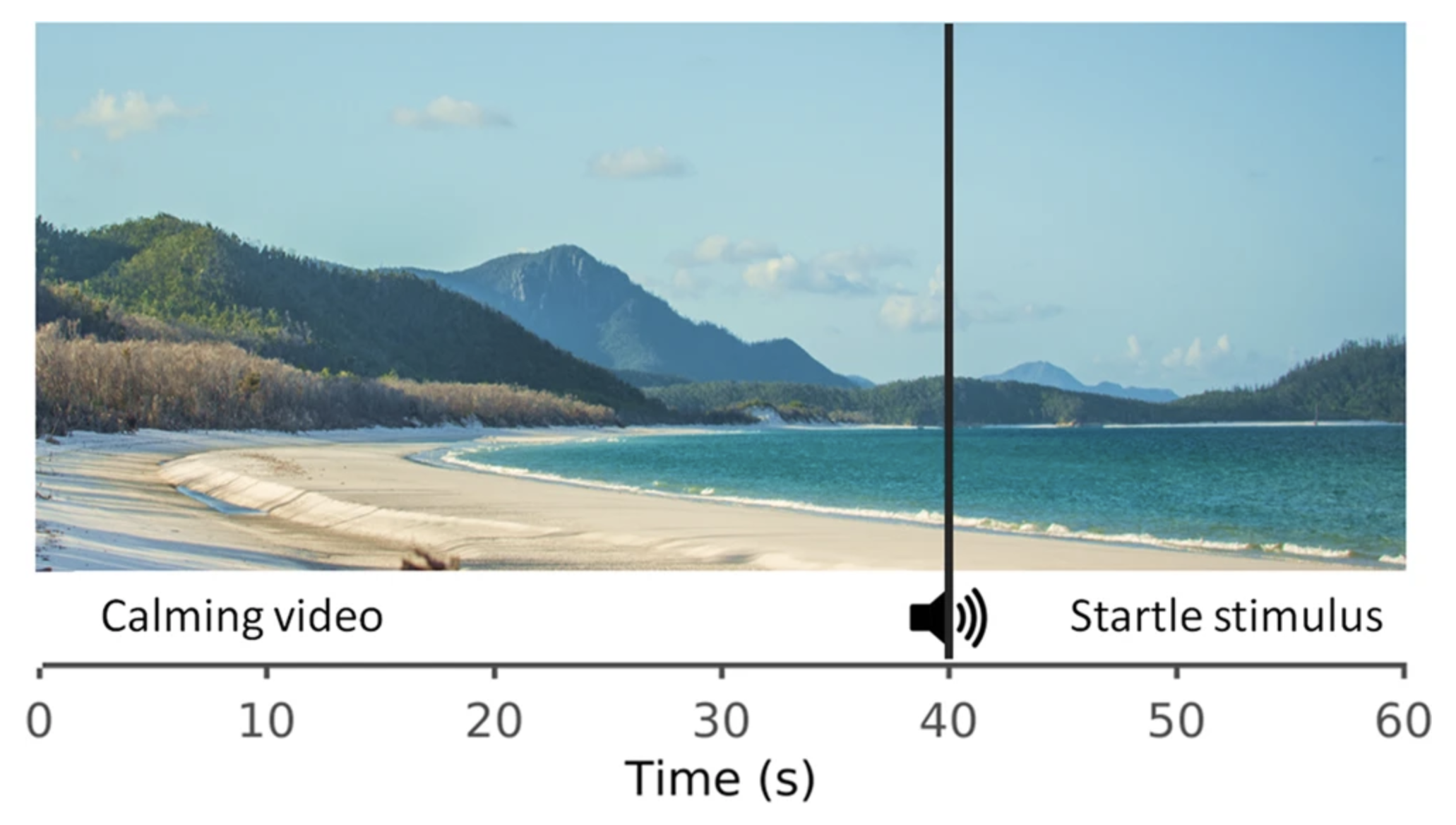}
\caption{Example of an emotional stimulus by Sonkusare et al. to elicit a spontaneous response. A calming ocean video clip was played for 60 seconds. A loud gunshot sound (80dB) was played at 40seconds to mimic a startle response. \citep{sonkusare2019detecting}}
\label{ocean}
\end{figure}

Another argument against discrete labels is the possibility that people express emotions as internalizers or externalizers, meaning different people suppress emotional expression in different ways making it difficult to truly capture expression in a basic, discrete manner \citep{gross1993emotional}. To elicit spontaneous response, emotion researchers use static images such as the International Affective Picture System \citep{kosonogov2017facial} or short clips of emotional videos \citep{nguyen2013thermal}. In a recent 2019 study by \citet{sonkusare2019detecting}, they use an auditory stimulus described in \autoref{ocean} to mimic a startle response, spontaneously.

\subsection{Provide social or personal context}
\begin{figure}[ht]
\centering
\includegraphics[width=0.35\textwidth]{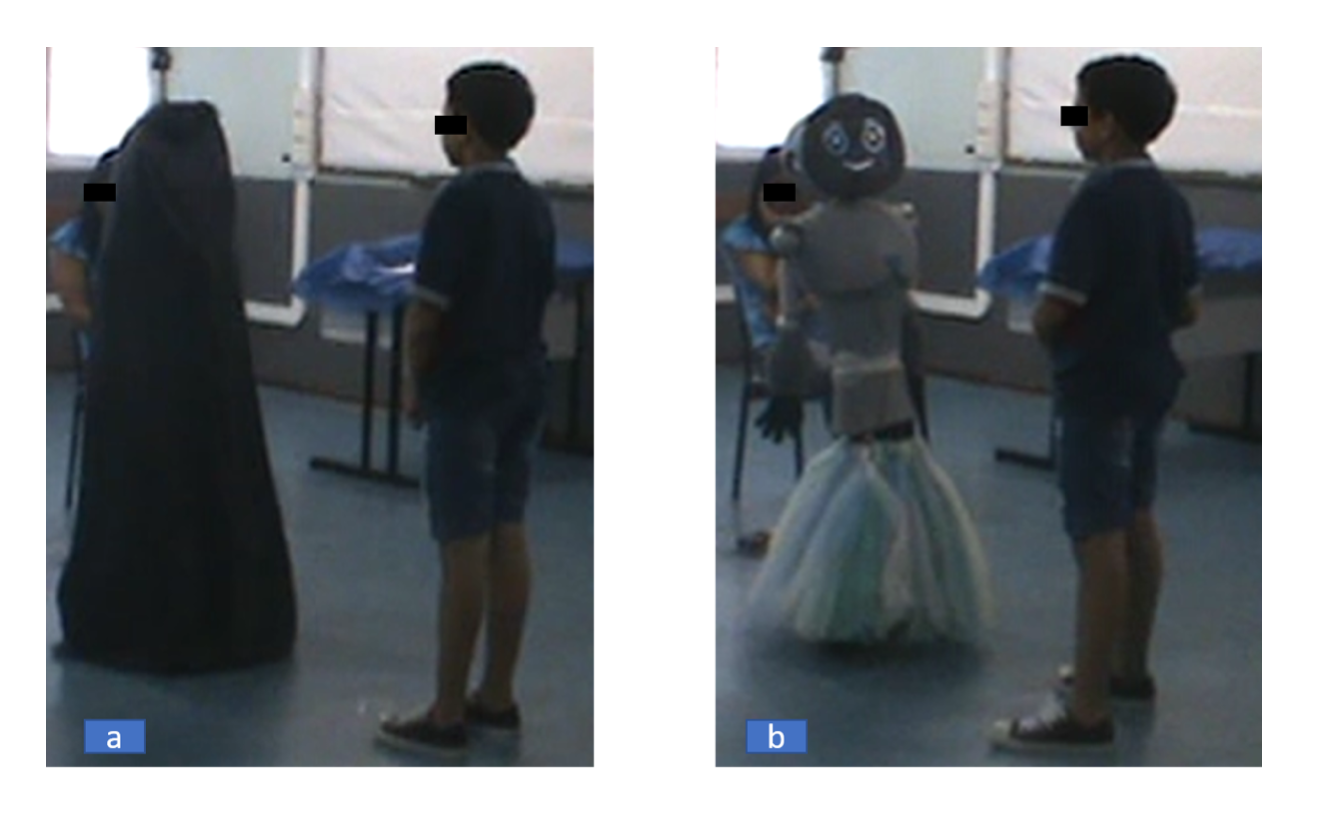}
\caption{Experimental setup showing the child-robot interaction by \citet{goulart2019emotion} (a) Before showing the robot; (b) After presenting it.}
\label{goulart}
\end{figure}

In a similar vein to spontaneous, natural emotion collection, providing social context in an experimental setting will change the nature of the emotion recorded. Context labeling to account for elicitation methods that are prompted spontaneously through personal elicitation (i.e. images, videos), versus social interaction with another person (or robot per \citep{goulart2019emotion}) may signal different physiological responses reflected in thermal imagery. Factors that influence these responses may include interpersonal distance, gaze direction, and opposite gender in the interaction \citep{kosonogov2017facial,gunes2010automatic}.  A sociodynamic model of emotions~\citep{mesquita2014emotions} asserts that emotions "emerge in interplay with and derive their specific function from the social context. This means that emotional experience and behavior will be differently constructed across various contexts". For example, Goulart \citep{goulart2019emotion} analyzed emotional response for 17 children during a human-child robot interaction experiment shown in \autoref{goulart}. Using Principle Component Analysis (PCA) and Linear Discriminant Analysis (LDA), they inferred happiness and surprise as the most frequently expressed, which were consistent with what the children self-reported upon interacting with the New-Mobile Autonomous Robot for Interaction with Autistics (N-MARIA) robot.  

\subsection{Collect multimodal pairs}
In 2000 Yoshitomi~\citep{yoshitomi2000effect} classified discrete affects by combining visible, thermal, and audio signals from 21 test subjects, achieving 85\% accuracy. \citet{zhu2007forehead} discussed multimodal data as "cross scale" data for biomedical research, or interconnections of different types of data using AI to infer mappings even if some data is missing. In essence, both were developing multimodal machine learning models, where multiple modalities, or types of information, may be combined to increase the accuracy of models \citep{baltruvsaitis2018multimodal}. The approach to collect pairs is not new. Nguyen collected thermal FER pairs for the KTFE database \citep{nguyen2013thermal} and the Iris \citep{iris}, Eurecom \citep{mallat2018benchmark}, and University of Notre Dame \citep{und} also have pairs which offer greater flexibility for different AI use cases like image translation for person re-identification. This includes research into thermal-to-visible GANs~\citep{mallat2018benchmark,kniaz2018thermalgan,chen2019matching,zhang2018tv}. With paired images capturing the RGB and LWIR images simultaneously using a camera equipped with a dual sensor, offers a mapping between both modalities for an AI algorithm to learn. 

\begin{figure}[ht]
\centering
\includegraphics[width=0.30\textwidth]{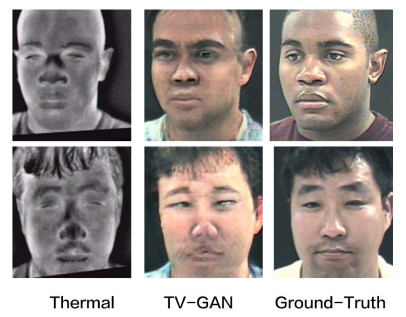}
\caption{Example of TV-GAN trained on multimodal pairs for thermal-to-visible image translation \citep{zhang2018tv}}
\label{tv-gan}
\end{figure}

\subsection{Document experimental setup}
Documenting experimental setup is important in order to minimize bias in the resulting thermogram, which can be affected by a variety of environmental and human subject conditions. Ioannao~ \citep{ioannou2014thermal} articulates in his paper on the potential and limitations of thermal imaging in physiology that, "Cutaneous thermal responses to external stimuli of psychophysiological valence could result in small temperature variations of the ROIs. Thus, it is extremely important to ensure that the observed temperature variations are not artifacts due to either environmental physiological causes or simply subject motion." Some of these can be minimized, the methods of which should be recorded and shared in the paper so that other thermal FER data collection trials can be repeated or improved to control for these external factors. 

\begin{figure}[ht]
\centering
\includegraphics[width=0.40\textwidth]{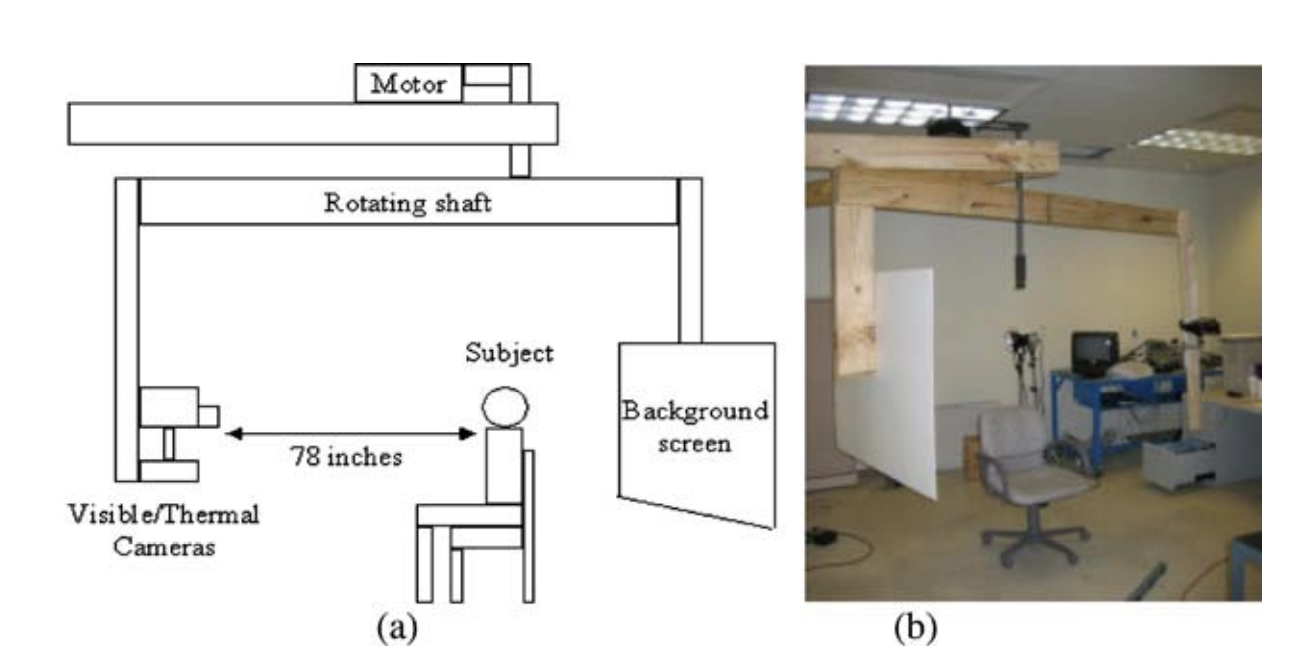}
\caption{Experimental Setup for Iris dataset capture \citep{kong2007multiscale}}
\label{kong}
\end{figure}

\begin{figure}[ht]
\centering
\includegraphics[width=\columnwidth]{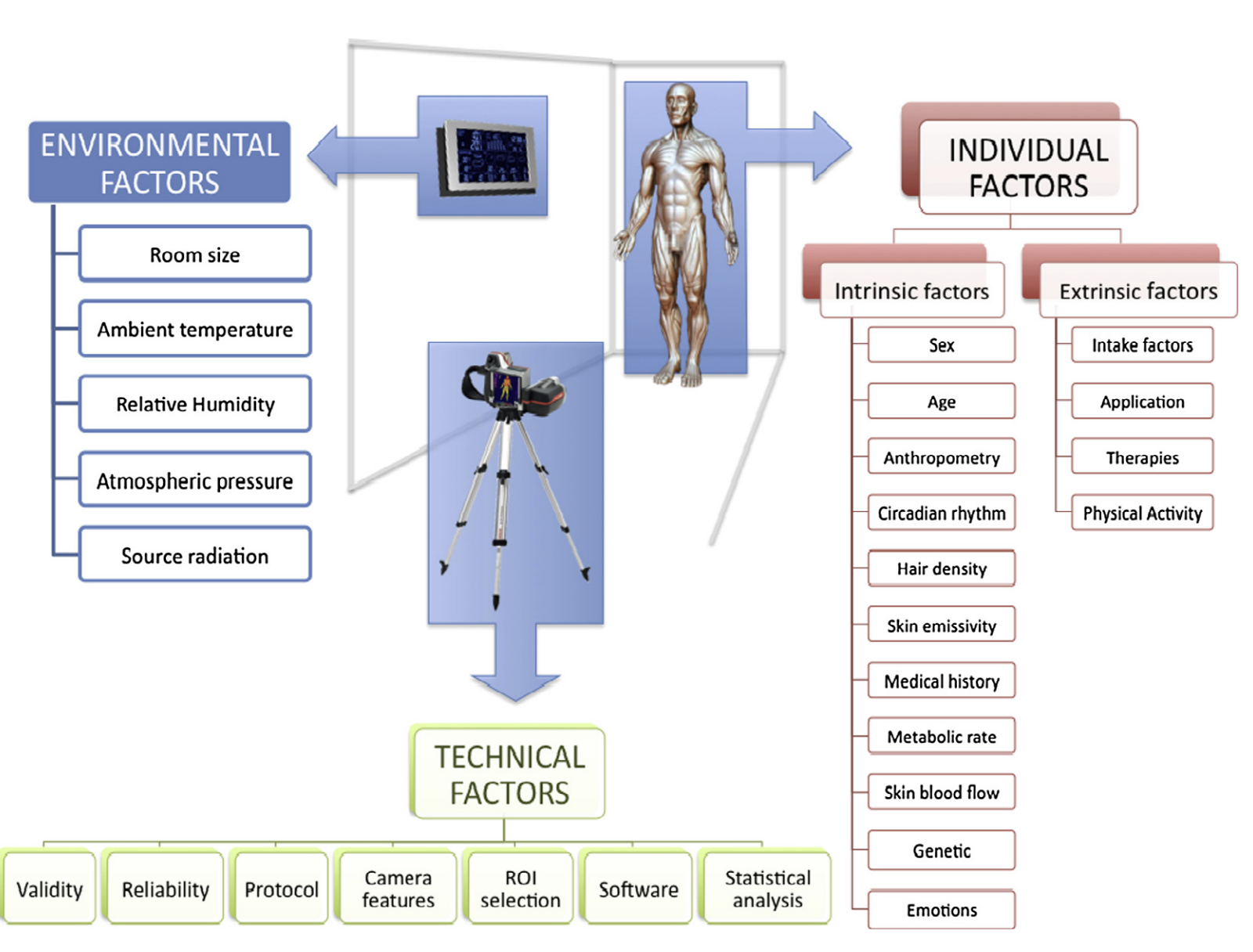}
\caption{Factors influencing thermal imagery of humans \citep{fernandez2015classification}}
\label{factors}
\end{figure}

In \autoref{design} we provide a sample of experimental parameters from several thermal FER papers and show how they vary from paper to paper. This demonstrates non-standard setups over the years of thermal FER research that could affect the reusability and generalization of these data for AI experiments. But, different papers vary in the extent of how much they document their experimental protocol provided in an example set of papers in \autoref{design}. Multiple factors need to be managed in order to minimize variables in the environment that influence thermal capture, leading to potentially misleading thermograms such as 1) Cold or warm air, as well as humidity, 2) Facial expressions (e.g. open mouth), 3) Physical conditions (e.g. lack of sleep, alcohol, caffeine), 4) Mental state (i.e. fear, stress, excitement), 5) Opaque to glasses, 6) Skin temperature variance through the day \citep{kosonogov2017facial}.  Fernandez et al. provide a comprehensive review of environmental, individual, and technical factors that influence IR reliability per \autoref{factors}~ \citep{fernandez2015classification}. 

Experimental design also includes the demographics of recruited subjects. Very few details are provided about race and ethnicity shown in \autoref{design} for the exception of \citep{lopez2017detecting} who indicated that nine out of 19 individuals were of Chinese ethnicity. With the ethical problems of visible FR in failing to train algorithms on a representative and balanced minority dataset, thermal FER researchers need to understand exactly what subjects are being included in the data and what underlying assumptions are being broadcast into training.  Further, we have so far been discussing thermal FER on adults in the various papers introduced. Very few studies, limited to \citep{goulart2019emotion} for child-robot interaction, \citep{ioannou2013autonomic} for guilt, \citep{ebisch2012mother} for child-mother imprinting, \citep{manini2013mom} for mother-child of vicarious autonomic response, collect thermal FER data on children. For the exception of Panetta et al., none of the thermal databases we identified appear to include children in their dataset, to the author's knowledge for thermal FER. So far, much work is still needed to generate an ethnically and age-diverse thermal FER dataset. Lastly, experimental set-up should also document technical methods that aim at normalizing the detected thermal face. For example, \citet{wang2014emotion} describes using the Otsu threshold algorithm to binarize the thermal images, detecting the face boundary, and removing baseline temperature to minimize the effects of temperature changes in the environment. Similar methods were introduced by Friedrich and Yeshurun in 2002 \citep{friedrich2002seeing}.

\begin{table*}[]
\caption{Summary of Thermal FER Data Challenge}
\label{summary}
\renewcommand{\arraystretch}{1.2}
\adjustbox{max width=\textwidth}{
\begin{tabular}{|p{5cm}p{5cm}p{5cm}p{5cm}|}
\hline
\textbf{Challenge} &
\textbf{Consequence} &
\textbf{Mitigation} &
\textbf{Opportunities} \\
  \hline
Include video sequences &
  Static images fail to capture the complete temporal dynamics of emotional   response. &
  Including labeled videos in thermal FER dataset. &
  Spatio-temporal labeling of thermal onset, delay, duration of   physiological response. \\
  \hline
Enable spontaneous response &
  Discrete posed expressions may not invoke realistic physiological   response. &
  Add spontaneous elicitation where possible, in addition to discrete set. &
  Natural, "in the wild" expressions that offer accurate   representations of emotion. \\
  \hline
Provide social or personal context &
  Thermal data collected without social stimuli may not be useable for   social use cases. &
  If appropriate, label social context or if controlling for, document how   social response has been minimized. &
  Social interaction thermal FER expressions, with labeled context and   scenarios. \\
  \hline
Collect multimodal pairs &
  No opportunity to increase accuracy or learn from additional modality mappings if only one modality (thermal) is collected. &
  May require dual sensor, or experimental design for simultaneous capture   using two cameras. &
  Multimodal pairs for various social, spontaneous elicited thermal FER   domains. \\
  \hline
Document experimental setup &
  Confounding through uncontrolled environmental variables can lead to   misleading images. &
  Report at minimum, the parameters shown in in Table 3. &
  Standard thermal FER experimental protocol for design and demographic   documentation. \\
  \hline
Accounting for Sensor Differences &
  Untested margin of error for images collected using different thermal   sensors. &
  No mitigation strategy. This is an open research question. &
  Assessment with optical engineers to determine margin of error across   sensors for human thermal FER.\\
\hline
\end{tabular}}
\end{table*}

\subsection{Accounting for Sensor Differences}
Lastly, the cost of thermal sensors through vendors like FLIR, have decreased over the past decade with increasingly higher quality resolution made accessible to the public. Prior papers have extensively used the Iris and Equinox (now discontinued) datasets. But with the release of more custom datasets as shown in \autoref{datasets}, is it fair to compare the output of thermal images from one sensor against another, which may have different optical properties? Or, is it sufficient that each sensor operates in the LWIR band? Many researchers have used different thermal sensors over the years: Pavlidis detected anxiety in thermal imagery in 2000 using an uncooled thermal camera with a spectral band of 8$\mu$m-14$\mu$m manufactured by Raytheon (the ExplorIR model) \citep{pavlidis2000imaging}, Nguyen in 2014 used a NEC R300 collecting in the 8$\mu$m-14$\mu$m  band \citep{nguyen2013thermal}, Aureli in 2015 used a FLIR SC660, an uncooled microbolometer sensor that collects in the 7.5$\mu$m – 13$\mu$m  band \citep{aureli2015behavioral}, and Eurecom researchers in 2018 used a FLIR Duo-Pro, an uncooled VOx Microbolometer sensor operating in 7.5$\mu$m–13.5 $\mu$m \citep{mallat2018benchmark}.  \autoref{design} provides a selection of thermal cameras used across various thermal FER studies as examples of how the cameras vary from study to study.

\section{Recommendations}
It is daunting to attempt to design a universal, thermal FER benchmark dataset that can account for the myriad of challenges we described. Extensive funding for time, labor, and evaluation would be required.  Some challenges are easier to mitigate than others, for example improving the documentation of experimental setup possibly using templates by  \citet{gebru2018datasheets} and  \citet{mitchell2019model} versus designing physiological stimuli. But, there may be more feasible short-term solutions that emphasize quality of reviewing the limitations of individual datasets and annotating each with a new labeling system. First, we have observed there are a number of custom datasets as described in \autoref{aithermalpapers} and are confident that our review missed several proprietary, unpublished, non-English, or classified thermal FER datasets. As a result, there are likely multiple thermal FER databases available all collected with a different set of subjects, experimental setups, and labeling. Offering these in a central online location, would be one step towards inventorying the breadth of data already available worldwide. 

\begin{figure}[ht]
\centering
\includegraphics[width=0.20\textwidth]{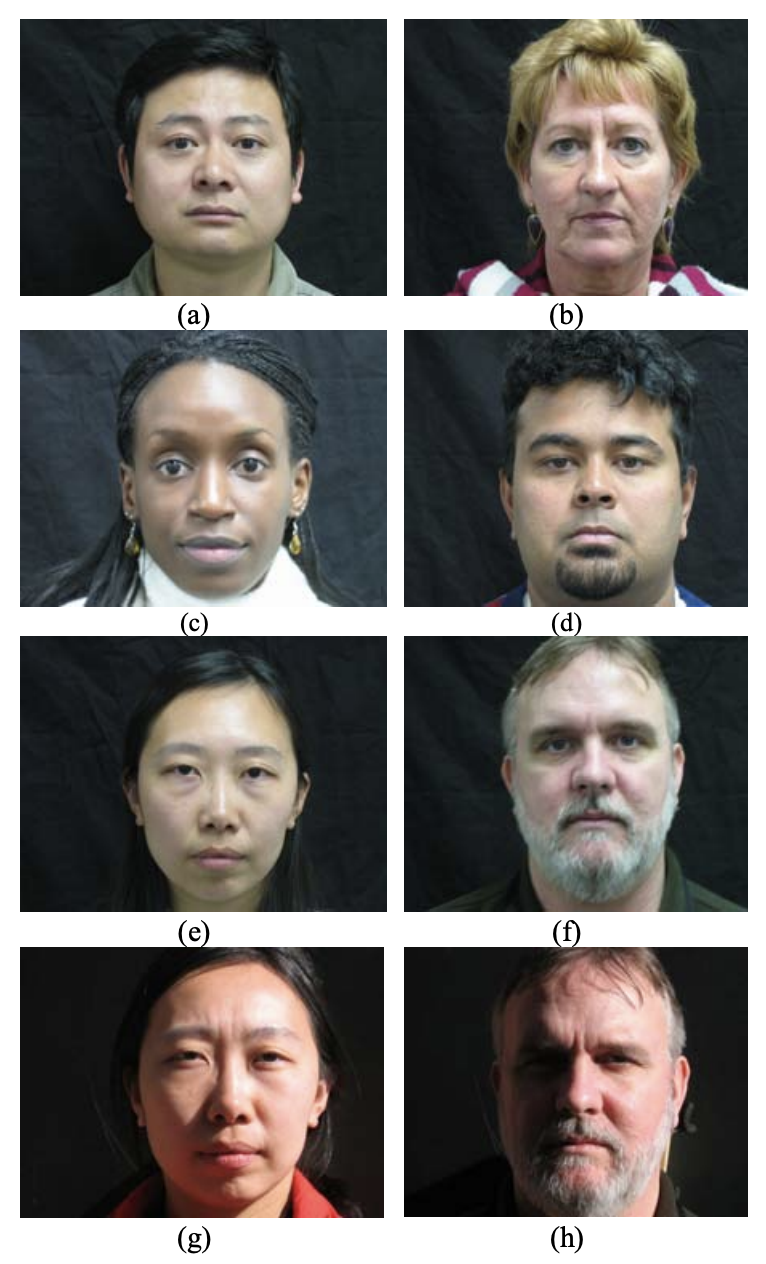}
\caption{Participants from diverse multimodal dataset collected by the IRIS Lab in 2006 \citep{chang2006indoor}}
\label{iris_lab}
\end{figure}

Secondly, combining across multiple existing thermal FER datasets and labeling by sensor, domain, posed or spontaneous emotion, resolution, and presence of social context, and stimulus, may be one step towards the aggregation of a larger database. Gathering training data across different datasets is not unusual in thermal FER, as previously noted when \citet{wang2014emotion} combined the NVIE and Equinox datasets to train their DBM model. Both first and second steps would require an effort across researchers to offer up and make available their thermal FER datasets.

Third, despite our review of the thermal FR and FER literature, we struggled to identify any research to evaluate the limits of obfuscating age, gender, ethnicity, and race using thermal imagery. Although some papers affirmed that their dataset consisted of diverse demographics \citep{chang2006indoor} per \autoref{iris_lab}, none to our knowledge, conducted quantitative tests with human reviewers and inter-rater statistics to test whether or not sensitive demographics could be masked. We believe that in order to assert that thermal imagery can afford any privacy protection and minimize bias, tests must be developed using IRB approval. More broadly, future work should take careful consideration into the scientific questions their research is tackling and the impact it may have in developing or prolonging undesired biases \citep{10.1145/230538.230561}. Biometrics related research is inherently sensitive and solutions can be valuable to society \citep{Jain2016a}. As such researchers should make sure they are familiar with ethical concerns that have occurred in neighboring application areas \citep{pmlr-v81-ensign18a,Chouldechova2017,Kleinberg2016} and remain open to understanding new perspective in which their research may be helpful or detrimental, and could be improved to reduce potential risks ~\citep{Skirpan2017,Goldsmith2017,applied_fairness_2018}. 

\section{Conclusion}
In this paper, we introduced the advantages of using thermal imagery over RGB for facial FER and provided a survey of thermal FER AI papers, datasets, and selected samples of experimental design protocols. There are several technical benefits of using thermal imagery compared to RGB images for FER, one of which potentially being semi-anonymity. However, there are few labeled, standard thermal affective data sets available for AI training. We have provided a summary of the proposed challenges, with our insights on the consequences, mitigation, and opportunities for each in \autoref{summary}. 

\section*{Acknowledgments}
We thank the three anonymous reviewers for AAAI 2020 for their feedback and comments. We also thank Steve Escaravage from Booz Allen Hamilton for his review of this article. This work is supported by grant CRII (IIS--1948399) from the National Science Foundation.

\bibliographystyle{aaai} 

\small{
\bibliography{main}

\begin{thebibliography}{}

\bibitem[\protect\citeauthoryear{Aureli \bgroup et al\mbox.\egroup
  }{2015}]{aureli2015behavioral}
Aureli, T.; Grazia, A.; Cardone, D.; and Merla, A.
\newblock 2015.
\newblock Behavioral and facial thermal variations in 3-to 4-month-old infants
  during the still-face paradigm.
\newblock {\em Frontiers in psychology} 6:1586.

\bibitem[\protect\citeauthoryear{Baltru{\v{s}}aitis, Ahuja, and
  Morency}{2018}]{baltruvsaitis2018multimodal}
Baltru{\v{s}}aitis, T.; Ahuja, C.; and Morency, L.-P.
\newblock 2018.
\newblock Multimodal machine learning: A survey and taxonomy.
\newblock {\em IEEE transactions on pattern analysis and machine intelligence}
  41(2):423--443.

\bibitem[\protect\citeauthoryear{Bourlai \bgroup et al\mbox.\egroup
  }{2012}]{bourlai2012use}
Bourlai, T.; Pryor, R.~R.; Suyama, J.; Reis, S.~E.; and Hostler, D.
\newblock 2012.
\newblock Use of thermal imagery for estimation of core body temperature during
  precooling, exertion, and recovery in wildland firefighter protective
  clothing.
\newblock {\em Prehospital Emergency Care}.

\bibitem[\protect\citeauthoryear{Buddharaju \bgroup et al\mbox.\egroup
  }{2007}]{buddharaju2007physiology}
Buddharaju, P.; Pavlidis, I.~T.; Tsiamyrtzis, P.; and Bazakos, M.
\newblock 2007.
\newblock Physiology-based face recognition in the thermal infrared spectrum.
\newblock {\em IEEE transactions on pattern analysis and machine intelligence}
  29(4):613--626.

\bibitem[\protect\citeauthoryear{Buolamwini and
  Gebru}{2018}]{buolamwini2018gender}
Buolamwini, J., and Gebru, T.
\newblock 2018.
\newblock Gender shades: Intersectional accuracy disparities in commercial
  gender classification.
\newblock In {\em Conference on fairness, accountability and transparency},
  77--91.

\bibitem[\protect\citeauthoryear{Chang \bgroup et al\mbox.\egroup
  }{2006}]{chang2006indoor}
Chang, H.; Harishwaran, H.; Yi, M.; Koschan, A.; Abidi, B.; and Abidi, M.
\newblock 2006.
\newblock An indoor and outdoor, multimodal, multispectral and multi-illuminant
  database for face recognition.
\newblock In {\em 2006 Conference on Computer Vision and Pattern Recognition
  Workshop (CVPRW'06)},  54--54.
\newblock IEEE.

\bibitem[\protect\citeauthoryear{Chen and Ross}{2019}]{chen2019matching}
Chen, C., and Ross, A.
\newblock 2019.
\newblock Matching thermal to visible face images using a semantic-guided
  generative adversarial network.
\newblock In {\em 2019 14th IEEE International Conference on Automatic Face \&
  Gesture Recognition (FG 2019)},  1--8.
\newblock IEEE.

\bibitem[\protect\citeauthoryear{Chouldechova}{2017}]{Chouldechova2017}
Chouldechova, A.
\newblock 2017.
\newblock {Fair prediction with disparate impact: A study of bias in recidivism
  prediction instruments}.
\newblock In {\em FAT ML Workshop}.

\bibitem[\protect\citeauthoryear{Ebisch \bgroup et al\mbox.\egroup
  }{2012}]{ebisch2012mother}
Ebisch, S.~J.; Aureli, T.; Bafunno, D.; Cardone, D.; Romani, G.~L.; and Merla,
  A.
\newblock 2012.
\newblock Mother and child in synchrony: thermal facial imprints of autonomic
  contagion.
\newblock {\em Biological psychology}.

\bibitem[\protect\citeauthoryear{Ekman}{1999}]{ekman1999basic}
Ekman, P.
\newblock 1999.
\newblock Basic emotions.
\newblock {\em Handbook of cognition and emotion} 98(45-60):16.

\bibitem[\protect\citeauthoryear{Ensign \bgroup et al\mbox.\egroup
  }{2018}]{pmlr-v81-ensign18a}
Ensign, D.; Friedler, S.~A.; Neville, S.; Scheidegger, C.; and
  Venkatasubramanian, S.
\newblock 2018.
\newblock {Runaway Feedback Loops in Predictive Policing}.
\newblock In Friedler, S.~A., and Wilson, C., eds., {\em Proceedings of the 1st
  Conference on Fairness, Accountability and Transparency}, volume~81 of {\em
  Proceedings of Machine Learning Research},  160--171.
\newblock New York, NY, USA: PMLR.

\bibitem[\protect\citeauthoryear{Equinox}{}]{equinox}
Equinox.
\newblock Equinox database at http://www.equinoxsensors.com/products/hid.html.

\bibitem[\protect\citeauthoryear{Fern{\'a}ndez-Cuevas \bgroup et
  al\mbox.\egroup }{2015}]{fernandez2015classification}
Fern{\'a}ndez-Cuevas, I.; Marins, J. C.~B.; Lastras, J.~A.; Carmona, P. M.~G.;
  Cano, S.~P.; Garc{\'\i}a-Concepci{\'o}n, M.~{\'A}.; and Sillero-Quintana, M.
\newblock 2015.
\newblock Classification of factors influencing the use of infrared
  thermography in humans: A review.
\newblock {\em Infrared Physics \& Technology} 71:28--55.

\bibitem[\protect\citeauthoryear{Friedman and
  Nissenbaum}{1996}]{10.1145/230538.230561}
Friedman, B., and Nissenbaum, H.
\newblock 1996.
\newblock {Bias in Computer Systems}.
\newblock {\em ACM Trans. Inf. Syst.} 14(3):330--347.

\bibitem[\protect\citeauthoryear{Friedrich and
  Yeshurun}{2002}]{friedrich2002seeing}
Friedrich, G., and Yeshurun, Y.
\newblock 2002.
\newblock Seeing people in the dark: Face recognition in infrared images.
\newblock In {\em International Workshop on Biologically Motivated Computer
  Vision},  348--359.
\newblock Springer.

\bibitem[\protect\citeauthoryear{Garbey \bgroup et al\mbox.\egroup
  }{2007}]{garbey2007contact}
Garbey, M.; Sun, N.; Merla, A.; and Pavlidis, I.
\newblock 2007.
\newblock Contact-free measurement of cardiac pulse based on the analysis of
  thermal imagery.
\newblock {\em IEEE transactions on Biomedical Engineering} 54(8).

\bibitem[\protect\citeauthoryear{Gebru \bgroup et al\mbox.\egroup
  }{2018}]{gebru2018datasheets}
Gebru, T.; Morgenstern, J.; Vecchione, B.; Vaughan, J.~W.; Wallach, H.;
  Daume{\'e}~III, H.; and Crawford, K.
\newblock 2018.
\newblock Datasheets for datasets.
\newblock {\em arXiv preprint arXiv:1803.09010}.

\bibitem[\protect\citeauthoryear{Ghiass \bgroup et al\mbox.\egroup
  }{2014}]{ghiass2014infrared}
Ghiass, R.~S.; Arandjelovi{\'c}, O.; Bendada, A.; and Maldague, X.
\newblock 2014.
\newblock Infrared face recognition: A comprehensive review of methodologies
  and databases.
\newblock {\em Pattern Recognition} 47(9):2807--2824.

\bibitem[\protect\citeauthoryear{Goldsmith and Burton}{2017}]{Goldsmith2017}
Goldsmith, J., and Burton, E.
\newblock 2017.
\newblock {Why Teaching Ethics to AI Practitioners Is Important}.
\newblock In {\em The AAAI-17 workshop on AI, Ethics, and Society},  110--114.

\bibitem[\protect\citeauthoryear{Goulart \bgroup et al\mbox.\egroup
  }{2019}]{goulart2019emotion}
Goulart, C.; Valad{\~a}o, C.; Delisle-Rodriguez, D.; Caldeira, E.; and Bastos,
  T.
\newblock 2019.
\newblock Emotion analysis in children through facial emissivity of infrared
  thermal imaging.
\newblock {\em PloS one} 14(3):e0212928.

\bibitem[\protect\citeauthoryear{Greene, Hoffmann, and
  Stark}{2019}]{greene2019better}
Greene, D.; Hoffmann, A.~L.; and Stark, L.
\newblock 2019.
\newblock Better, nicer, clearer, fairer: A critical assessment of the movement
  for ethical artificial intelligence and machine learning.
\newblock In {\em Proceedings of the 52nd Hawaii International Conference on
  System Sciences}.

\bibitem[\protect\citeauthoryear{Grgic}{}]{scface}
Grgic, M.
\newblock Scface - surveillance cameras face database.

\bibitem[\protect\citeauthoryear{Gross and Levenson}{1993}]{gross1993emotional}
Gross, J.~J., and Levenson, R.~W.
\newblock 1993.
\newblock Emotional suppression: physiology, self-report, and expressive
  behavior.
\newblock {\em Journal of personality and social psychology} 64(6):970.

\bibitem[\protect\citeauthoryear{Gunes and Pantic}{2010}]{gunes2010automatic}
Gunes, H., and Pantic, M.
\newblock 2010.
\newblock Automatic, dimensional and continuous emotion recognition.
\newblock {\em International Journal of Synthetic Emotions (IJSE)}.

\bibitem[\protect\citeauthoryear{Hahn \bgroup et al\mbox.\egroup
  }{2012}]{hahn2012hot}
Hahn, A.~C.; Whitehead, R.~D.; Albrecht, M.; Lefevre, C.~E.; and Perrett, D.~I.
\newblock 2012.
\newblock Hot or not? thermal reactions to social contact.
\newblock {\em Biology letters} 8(5):864--867.

\bibitem[\protect\citeauthoryear{Hamdi}{2020}]{hamdi_2020}
Hamdi, T.
\newblock 2020.
\newblock Deepfake detection challenge.

\bibitem[\protect\citeauthoryear{Hammoud}{}]{iris}
Hammoud, R.~I.
\newblock Otcbvs benchmark dataset collection.

\bibitem[\protect\citeauthoryear{Heo \bgroup et al\mbox.\egroup
  }{2004}]{heo2004fusion}
Heo, J.; Kong, S.~G.; Abidi, B.~R.; and Abidi, M.~A.
\newblock 2004.
\newblock Fusion of visual and thermal signatures with eyeglass removal for
  robust face recognition.
\newblock In {\em 2004 Conference on Computer Vision and Pattern Recognition
  Workshop},  122--122.
\newblock IEEE.

\bibitem[\protect\citeauthoryear{Hermosilla \bgroup et al\mbox.\egroup
  }{2012}]{hermosilla2012comparative}
Hermosilla, G.; Ruiz-del Solar, J.; Verschae, R.; and Correa, M.
\newblock 2012.
\newblock A comparative study of thermal face recognition methods in
  unconstrained environments.
\newblock {\em Pattern Recognition} 45(7):2445--2459.

\bibitem[\protect\citeauthoryear{Ioannou \bgroup et al\mbox.\egroup
  }{2013}]{ioannou2013autonomic}
Ioannou, S.; Ebisch, S.; Aureli, T.; Bafunno, D.; Ioannides, H.~A.; Cardone,
  D.; Manini, B.; Romani, G.~L.; Gallese, V.; and Merla, A.
\newblock 2013.
\newblock The autonomic signature of guilt in children: a thermal infrared
  imaging study.
\newblock {\em PloS one} 8(11):e79440.

\bibitem[\protect\citeauthoryear{Ioannou, Gallese, and
  Merla}{2014}]{ioannou2014thermal}
Ioannou, S.; Gallese, V.; and Merla, A.
\newblock 2014.
\newblock Thermal infrared imaging in psychophysiology: potentialities and
  limits.
\newblock {\em Psychophysiology} 51(10):951--963.

\bibitem[\protect\citeauthoryear{Jai}{2016}]{Jain2016a}
2016.
\newblock {50 years of biometric research: Accomplishments, challenges, and
  opportunities}.

\bibitem[\protect\citeauthoryear{Jarlier \bgroup et al\mbox.\egroup
  }{2011}]{jarlier2011thermal}
Jarlier, S.; Grandjean, D.; Delplanque, S.; N'diaye, K.; Cayeux, I.; Velazco,
  M.~I.; Sander, D.; Vuilleumier, P.; and Scherer, K.~R.
\newblock 2011.
\newblock Thermal analysis of facial muscles contractions.
\newblock {\em IEEE transactions on affective computing} 2(1):2--9.

\bibitem[\protect\citeauthoryear{Khan, Ingleby, and
  Ward}{2006}]{khan2006automated}
Khan, M.~M.; Ingleby, M.; and Ward, R.~D.
\newblock 2006.
\newblock Automated facial expression classification and affect interpretation
  using infrared measurement of facial skin temperature variations.
\newblock {\em ACM Transactions on Autonomous and Adaptive Systems (TAAS)}
  1(1):91--113.

\bibitem[\protect\citeauthoryear{Kleinberg, Mullainathan, and
  Raghavan}{2016}]{Kleinberg2016}
Kleinberg, J.; Mullainathan, S.; and Raghavan, M.
\newblock 2016.
\newblock {Inherent Trade-Offs in the Fair Determination of Risk Scores}.
\newblock In {\em FAT ML Workshop}.

\bibitem[\protect\citeauthoryear{Kniaz \bgroup et al\mbox.\egroup
  }{2018}]{kniaz2018thermalgan}
Kniaz, V.~V.; Knyaz, V.~A.; Hladuvka, J.; Kropatsch, W.~G.; and Mizginov, V.
\newblock 2018.
\newblock Thermalgan: Multimodal color-to-thermal image translation for person
  re-identification in multispectral dataset.
\newblock In {\em ECCV}.

\bibitem[\protect\citeauthoryear{Kong \bgroup et al\mbox.\egroup
  }{2007}]{kong2007multiscale}
Kong, S.~G.; Heo, J.; Boughorbel, F.; Zheng, Y.; Abidi, B.~R.; Koschan, A.; Yi,
  M.; and Abidi, M.~A.
\newblock 2007.
\newblock Multiscale fusion of visible and thermal ir images for
  illumination-invariant face recognition.
\newblock {\em International Journal of Computer Vision} 71(2):215--233.

\bibitem[\protect\citeauthoryear{Kopaczka, Kolk, and
  Merhof}{2018}]{kopaczka2018fully}
Kopaczka, M.; Kolk, R.; and Merhof, D.
\newblock 2018.
\newblock A fully annotated thermal face database and its application for
  thermal facial expression recognition.
\newblock In {\em 2018 IEEE International Instrumentation and Measurement
  Technology Conference (I2MTC)},  1--6.
\newblock IEEE.

\bibitem[\protect\citeauthoryear{Kosonogov \bgroup et al\mbox.\egroup
  }{2017}]{kosonogov2017facial}
Kosonogov, V.; De~Zorzi, L.; Honore, J.; Mart{\'\i}nez-Vel{\'a}zquez, E.~S.;
  Nandrino, J.-L.; Martinez-Selva, J.~M.; and Sequeira, H.
\newblock 2017.
\newblock Facial thermal variations: A new marker of emotional arousal.
\newblock {\em PloS one} 12(9):e0183592.

\bibitem[\protect\citeauthoryear{Kumar}{}]{iit}
Kumar, A.
\newblock Iit delhi near ir face database version 2.0.

\bibitem[\protect\citeauthoryear{Levenson}{1988}]{levenson1988emotion}
Levenson, R.~W.
\newblock 1988.
\newblock Emotion and the autonomic nervous system: A prospectus for research
  on autonomic specificity.
\newblock {\em Social psychophysiology: Theory and clinical applications}.

\bibitem[\protect\citeauthoryear{Li and Deng}{2018}]{li2018deep}
Li, S., and Deng, W.
\newblock 2018.
\newblock Deep facial expression recognition: A survey.
\newblock {\em arXiv preprint arXiv:1804.08348}.

\bibitem[\protect\citeauthoryear{Lohr}{2018}]{lohr_2018}
Lohr, S.
\newblock 2018.
\newblock Facial recognition is accurate, if you're a white guy.

\bibitem[\protect\citeauthoryear{Lopez, del Blanco, and
  Garcia}{2017}]{lopez2017detecting}
Lopez, M.~B.; del Blanco, C.~R.; and Garcia, N.
\newblock 2017.
\newblock Detecting exercise-induced fatigue using thermal imaging and deep
  learning.
\newblock In {\em 2017 Seventh International Conference on Image Processing
  Theory, Tools and Applications (IPTA)},  1--6.
\newblock IEEE.

\bibitem[\protect\citeauthoryear{Mallat and
  Dugelay}{2018}]{mallat2018benchmark}
Mallat, K., and Dugelay, J.-L.
\newblock 2018.
\newblock A benchmark database of visible and thermal paired face images across
  multiple variations.
\newblock In {\em 2018 BIOSIG},  1--5.
\newblock IEEE.

\bibitem[\protect\citeauthoryear{Mallat \bgroup et al\mbox.\egroup
  }{2019}]{mallat2019cross}
Mallat, K.; Damer, N.; Boutros, F.; Kuijper, A.; and Dugelay, J.-L.
\newblock 2019.
\newblock Cross-spectrum thermal to visible face recognition based on cascaded
  image synthesis.
\newblock In {\em 2019 International Conference on Biometrics (ICB)},  1--8.
\newblock IEEE.

\bibitem[\protect\citeauthoryear{Manini \bgroup et al\mbox.\egroup
  }{2013}]{manini2013mom}
Manini, B.; Cardone, D.; Ebisch, S.; Bafunno, D.; Aureli, T.; and Merla, A.
\newblock 2013.
\newblock Mom feels what her child feels: thermal signatures of vicarious
  autonomic response while watching children in a stressful situation.
\newblock {\em Frontiers in human neuroscience} 7:299.

\bibitem[\protect\citeauthoryear{Martinez-Martin}{2019}]{martinez2019important}
Martinez-Martin, N.
\newblock 2019.
\newblock What are important ethical implications of using facial recognition
  technology in health care?
\newblock {\em AMA journal of ethics} 21(2):E180.

\bibitem[\protect\citeauthoryear{Matsakis}{2020}]{matsakis_2020}
Matsakis, L.
\newblock 2020.
\newblock Amazon won't let police use its facial-recognition tech for one year.

\bibitem[\protect\citeauthoryear{McDuff, Girard, and
  El~Kaliouby}{2017}]{mcduff2017large}
McDuff, D.; Girard, J.~M.; and El~Kaliouby, R.
\newblock 2017.
\newblock Large-scale observational evidence of cross-cultural differences in
  facial behavior.
\newblock {\em Journal of Nonverbal Behavior} 41(1):1--19.

\bibitem[\protect\citeauthoryear{Merla \bgroup et al\mbox.\egroup
  }{2004}]{merla2004emotion}
Merla, A.; Di~Donato, L.; Rossini, P.; and Romani, G.
\newblock 2004.
\newblock Emotion detection through functional infrared imaging: preliminary
  results.
\newblock {\em Biomedizinische Technick} 48(2):284--286.

\bibitem[\protect\citeauthoryear{Merla}{2014}]{merla2014revealing}
Merla, A.
\newblock 2014.
\newblock Revealing psychophysiology and emotions through thermal infrared
  imaging.
\newblock In {\em PhyCS},  368--377.

\bibitem[\protect\citeauthoryear{Mesquita and
  Boiger}{2014}]{mesquita2014emotions}
Mesquita, B., and Boiger, M.
\newblock 2014.
\newblock Emotions in context: A sociodynamic model of emotions.
\newblock {\em Emotion Review} 6(4):298--302.

\bibitem[\protect\citeauthoryear{Mitchell \bgroup et al\mbox.\egroup
  }{2019}]{mitchell2019model}
Mitchell, M.; Wu, S.; Zaldivar, A.; Barnes, P.; Vasserman, L.; Hutchinson, B.;
  Spitzer, E.; Raji, I.~D.; and Gebru, T.
\newblock 2019.
\newblock Model cards for model reporting.
\newblock In {\em Proceedings of the conference on fairness, accountability,
  and transparency},  220--229.

\bibitem[\protect\citeauthoryear{Murgia}{2019}]{murgia_2019}
Murgia, M.
\newblock 2019.
\newblock Microsoft quietly deletes largest public face recognition data set.

\bibitem[\protect\citeauthoryear{Nguyen and Park}{2016}]{nguyen2016body}
Nguyen, D.~T., and Park, K.~R.
\newblock 2016.
\newblock Body-based gender recognition using images from visible and thermal
  cameras.
\newblock {\em Sensors} 16(2):156.

\bibitem[\protect\citeauthoryear{Nguyen \bgroup et al\mbox.\egroup
  }{2013}]{nguyen2013thermal}
Nguyen, H.; Kotani, K.; Chen, F.; and Le, B.
\newblock 2013.
\newblock A thermal facial emotion database and its analysis.
\newblock In {\em Pacific-Rim Symposium on Image and Video Technology}.
\newblock Springer.

\bibitem[\protect\citeauthoryear{Nhan and Chau}{2009}]{nhan2009classifying}
Nhan, B.~R., and Chau, T.
\newblock 2009.
\newblock Classifying affective states using thermal infrared imaging of the
  human face.
\newblock {\em IEEE Transactions on Biomedical Engineering} 57(4):979--987.

\bibitem[\protect\citeauthoryear{Panetta \bgroup et al\mbox.\egroup
  }{2018}]{panetta2018comprehensive}
Panetta, K.; Wan, Q.; Agaian, S.; Rajeev, S.; Kamath, S.; Rajendran, R.; Rao,
  S.; Kaszowska, A.; Taylor, H.; Samani, A.; et~al.
\newblock 2018.
\newblock A comprehensive database for benchmarking imaging systems.
\newblock {\em IEEE transactions on pattern analysis and machine intelligence}.

\bibitem[\protect\citeauthoryear{Pavlidis and
  Symosek}{2000}]{pavlidis2000imaging}
Pavlidis, I., and Symosek, P.
\newblock 2000.
\newblock The imaging issue in an automatic face/disguise detection system.
\newblock In {\em Proceedings IEEE Workshop on Computer Vision Beyond the
  Visible Spectrum: Methods and Applications (Cat. No. PR00640)},  15--24.
\newblock IEEE.

\bibitem[\protect\citeauthoryear{Pavlidis \bgroup et al\mbox.\egroup
  }{2007}]{pavlidis2007interacting}
Pavlidis, I.; Dowdall, J.; Sun, N.; Puri, C.; Fei, J.; and Garbey, M.
\newblock 2007.
\newblock Interacting with human physiology.
\newblock {\em Computer Vision and Image Understanding} 108(1-2):150--170.

\bibitem[\protect\citeauthoryear{Pavlidis \bgroup et al\mbox.\egroup
  }{2012}]{pavlidis2012fast}
Pavlidis, I.; Tsiamyrtzis, P.; Shastri, D.; Wesley, A.; Zhou, Y.; Lindner, P.;
  Buddharaju, P.; Joseph, R.; Mandapati, A.; Dunkin, B.; et~al.
\newblock 2012.
\newblock Fast by nature-how stress patterns define human experience and
  performance in dexterous tasks.
\newblock {\em Scientific Reports} 2:305.

\bibitem[\protect\citeauthoryear{Pittaluga, Zivkovic, and
  Koppal}{2016}]{pittaluga2016sensor}
Pittaluga, F.; Zivkovic, A.; and Koppal, S.~J.
\newblock 2016.
\newblock Sensor-level privacy for thermal cameras.
\newblock In {\em 2016 IEEE International Conference on Computational
  Photography (ICCP)},  1--12.
\newblock IEEE.

\bibitem[\protect\citeauthoryear{Puri \bgroup et al\mbox.\egroup
  }{2005}]{puri2005stresscam}
Puri, C.; Olson, L.; Pavlidis, I.; Levine, J.; and Starren, J.
\newblock 2005.
\newblock Stresscam: non-contact measurement of users' emotional states through
  thermal imaging.
\newblock In {\em CHI'05 extended abstracts on Human factors in computing
  systems}.

\bibitem[\protect\citeauthoryear{Raff}{2019}]{raff2019step}
Raff, E.
\newblock 2019.
\newblock A step toward quantifying independently reproducible machine learning
  research.
\newblock In {\em Advances in Neural Information Processing Systems},
  5485--5495.

\bibitem[\protect\citeauthoryear{Reid \bgroup et al\mbox.\egroup
  }{2013}]{reid2013soft}
Reid, D.~A.; Samangooei, S.; Chen, C.; Nixon, M.~S.; and Ross, A.
\newblock 2013.
\newblock Soft biometrics for surveillance: an overview.
\newblock In {\em Handbook of statistics}, volume~31. Elsevier.
\newblock  327--352.

\bibitem[\protect\citeauthoryear{Salazar-L{\'o}pez \bgroup et al\mbox.\egroup
  }{2015}]{salazar2015mental}
Salazar-L{\'o}pez, E.; Dom{\'\i}nguez, E.; Ramos, V.~J.; De~la Fuente, J.;
  Meins, A.; Iborra, O.; G{\'a}lvez, G.; Rodr{\'\i}guez-Artacho, M.; and
  G{\'o}mez-Mil{\'a}n, E.
\newblock 2015.
\newblock The mental and subjective skin: Emotion, empathy, feelings and
  thermography.
\newblock {\em Consciousness and cognition}.

\bibitem[\protect\citeauthoryear{Seo and Chung}{2019}]{seo2019face}
Seo, J., and Chung, I.-J.
\newblock 2019.
\newblock Face liveness detection using thermal face-cnn with external
  knowledge.
\newblock {\em Symmetry} 11(3):360.

\bibitem[\protect\citeauthoryear{Shoja~Ghiass}{2014}]{shoja2014face}
Shoja~Ghiass, R.
\newblock 2014.
\newblock Face recognition using infrared vision.

\bibitem[\protect\citeauthoryear{Sim{\'o}n \bgroup et al\mbox.\egroup
  }{2016}]{simon2016improved}
Sim{\'o}n, M.~O.; Corneanu, C.; Nasrollahi, K.; Nikisins, O.; Escalera, S.;
  Sun, Y.; Li, H.; Sun, Z.; Moeslund, T.~B.; and Greitans, M.
\newblock 2016.
\newblock Improved rgb-dt based face recognition.
\newblock {\em Iet Biometrics} 5(4):297--303.

\bibitem[\protect\citeauthoryear{Singer and Metz}{2019}]{singer_metz_2019}
Singer, N., and Metz, C.
\newblock 2019.
\newblock Many facial-recognition systems are biased, says u.s. study.

\bibitem[\protect\citeauthoryear{Skirpan and Gorelick}{2017}]{Skirpan2017}
Skirpan, M., and Gorelick, M.
\newblock 2017.
\newblock {The Authority of "Fair" in Machine Learning}.
\newblock In {\em FAT ML Workshop}.

\bibitem[\protect\citeauthoryear{Sonkusare \bgroup et al\mbox.\egroup
  }{2019}]{sonkusare2019detecting}
Sonkusare, S.; Ahmedt-Aristizabal, D.; Aburn, M.~J.; Nguyen, V.~T.; Pang, T.;
  Frydman, S.; Denman, S.; Fookes, C.; Breakspear, M.; and Guo, C.~C.
\newblock 2019.
\newblock Detecting changes in facial temperature induced by a sudden auditory
  stimulus based on deep learning-assisted face tracking.
\newblock {\em Scientific reports} 9(1):1--11.

\bibitem[\protect\citeauthoryear{Sylvester and
  Raff}{2018}]{applied_fairness_2018}
Sylvester, J., and Raff, E.
\newblock 2018.
\newblock {What About Applied Fairness?}
\newblock In {\em Machine Learning: The Debates (ML-D) organized as part of the
  Federated AI Meeting (FAIM 2018)}.

\bibitem[\protect\citeauthoryear{Tian, Kanade, and Cohn}{2005}]{tian2005facial}
Tian, Y.-L.; Kanade, T.; and Cohn, J.~F.
\newblock 2005.
\newblock Facial expression analysis.
\newblock In {\em Handbook of face recognition}. Springer.
\newblock  247--275.

\bibitem[\protect\citeauthoryear{Ting \bgroup et al\mbox.\egroup
  }{2020}]{ting2020digital}
Ting, D. S.~W.; Carin, L.; Dzau, V.; and Wong, T.~Y.
\newblock 2020.
\newblock Digital technology and covid-19.
\newblock {\em Nature medicine} 26(4):459--461.

\bibitem[\protect\citeauthoryear{Trujillo \bgroup et al\mbox.\egroup
  }{2005}]{Trujillo2005automatic}
Trujillo, L.; Olague, G.; Hammoud, R.; and Hernandez, B.
\newblock 2005.
\newblock Automatic feature localization in thermal images for facial
  expression recognition.
\newblock In {\em 2005 IEEE Computer Society Conference on Computer Vision and
  Pattern Recognition (CVPR'05)-Workshops}.
\newblock IEEE.

\bibitem[\protect\citeauthoryear{UND}{}]{und}
UND.
\newblock University of notre dame und-collection c.

\bibitem[\protect\citeauthoryear{Van~Natta \bgroup et al\mbox.\egroup
  }{2020}]{vannatta}
Van~Natta, M.; Chen, P.; Herbek, S.; Jain, R.; Kastelic, N.; Katz, E.; Struble,
  M.; Vanam, V.; and Vattikonda, N.
\newblock 2020.
\newblock rise and regulation of thermal facial recognition technology during
  the covid-19 pandemic.

\bibitem[\protect\citeauthoryear{Wang \bgroup et al\mbox.\egroup
  }{2010}]{wang2010natural}
Wang, S.; Liu, Z.; Lv, S.; Lv, Y.; Wu, G.; Peng, P.; Chen, F.; and Wang, X.
\newblock 2010.
\newblock A natural visible and infrared facial expression database for
  expression recognition and emotion inference.
\newblock {\em IEEE Transactions on Multimedia} 12(7):682--691.

\bibitem[\protect\citeauthoryear{Wang \bgroup et al\mbox.\egroup
  }{2014a}]{wang2014emotion}
Wang, S.; He, M.; Gao, Z.; He, S.; and Ji, Q.
\newblock 2014a.
\newblock Emotion recognition from thermal infrared images using deep boltzmann
  machine.
\newblock {\em Frontiers of Computer Science} 8(4):609--618.

\bibitem[\protect\citeauthoryear{Wang \bgroup et al\mbox.\egroup
  }{2014b}]{wang2014fusion}
Wang, S.; He, S.; Wu, Y.; He, M.; and Ji, Q.
\newblock 2014b.
\newblock Fusion of visible and thermal images for facial expression
  recognition.
\newblock {\em Frontiers of Computer Science} 8(2):232--242.

\bibitem[\protect\citeauthoryear{Wilder \bgroup et al\mbox.\egroup
  }{1996}]{wilder1996comparison}
Wilder, J.; Phillips, P.~J.; Jiang, C.; and Wiener, S.
\newblock 1996.
\newblock Comparison of visible and infra-red imagery for face recognition.
\newblock In {\em Proceedings of the Second International Conference on
  Automatic Face and Gesture Recognition},  182--187.
\newblock IEEE.

\bibitem[\protect\citeauthoryear{Yan \bgroup et al\mbox.\egroup
  }{2014}]{yan2014casme}
Yan, W.-J.; Li, X.; Wang, S.-J.; Zhao, G.; Liu, Y.-J.; Chen, Y.-H.; and Fu, X.
\newblock 2014.
\newblock Casme ii: An improved spontaneous micro-expression database and the
  baseline evaluation.
\newblock {\em PloS one} 9(1):e86041.

\bibitem[\protect\citeauthoryear{Yoshitomi \bgroup et al\mbox.\egroup
  }{2000}]{yoshitomi2000effect}
Yoshitomi, Y.; Kim, S.-I.; Kawano, T.; and Kilazoe, T.
\newblock 2000.
\newblock Effect of sensor fusion for recognition of emotional states using
  voice, face image and thermal image of face.
\newblock In {\em Proceedings 9th IEEE International Workshop on Robot and
  Human Interactive Communication. IEEE RO-MAN 2000 (Cat. No. 00TH8499)}.
\newblock IEEE.

\bibitem[\protect\citeauthoryear{Zhang \bgroup et al\mbox.\egroup
  }{2010}]{zhang2010directional}
Zhang, B.; Zhang, L.; Zhang, D.; and Shen, L.
\newblock 2010.
\newblock Directional binary code with application to polyu near-infrared face
  database.
\newblock {\em Pattern Recognition Letters} 31(14):2337--2344.

\bibitem[\protect\citeauthoryear{Zhang \bgroup et al\mbox.\egroup
  }{2018}]{zhang2018tv}
Zhang, T.; Wiliem, A.; Yang, S.; and Lovell, B.
\newblock 2018.
\newblock Tv-gan: Generative adversarial network based thermal to visible face
  recognition.
\newblock In {\em 2018 ICB}.
\newblock IEEE.

\bibitem[\protect\citeauthoryear{Zhu, Tsiamyrtzis, and
  Pavlidis}{2007}]{zhu2007forehead}
Zhu, Z.; Tsiamyrtzis, P.; and Pavlidis, I.
\newblock 2007.
\newblock Forehead thermal signature extraction in lie detection.
\newblock In {\em 2007 29th Annual International Conference of the IEEE
  Engineering in Medicine and Biology Society},  243--246.
\newblock IEEE.

\end{thebibliography}
}

\end{document}